\documentclass[11pt]{article}
\usepackage[a4paper,margin=1in]{geometry}
\usepackage[T1]{fontenc}
\usepackage[utf8]{inputenc}
\usepackage{lmodern}
\usepackage{microtype}
\usepackage{graphicx}
\usepackage{amsmath,amssymb}
\usepackage{booktabs}
\usepackage{multirow}
\usepackage{array}
\usepackage{tabularx}
\usepackage{longtable}
\usepackage{pdflscape}
\usepackage{placeins}
\usepackage{enumitem}
\usepackage{authblk}
\usepackage{xcolor}
\usepackage{hyperref}
\usepackage[numbers,sort]{natbib}
\hypersetup{
  colorlinks=true,
  linkcolor=blue,
  citecolor=blue,
  urlcolor=blue,
  pdftitle={From ECG Signals to Representative-Morphology Heatmaps for Biometric Recognition},
  pdfauthor={Athanasios Angelakis and Marta Gomez-Barrero},
  pdfsubject={ECG-derived image representations for biometric verification and identification},
  pdfkeywords={ECG biometrics, representative-morphology heatmaps, verification, identification, ZACH-ViT, visual transfer learning}
}
\newcommand{\R}{\mathbb{R}}
\newcommand{\fnmrten}{FNMR at FMR $\leq 10\%$}
\newcommand{\fnmrone}{FNMR at FMR $\leq 1\%$}
\title{From ECG Signals to Representative-Morphology Heatmaps for Biometric Recognition}
\author[1,2]{Athanasios Angelakis\thanks{ORCID: \href{https://orcid.org/0000-0003-1226-9560}{0000-0003-1226-9560}}}
\author[1]{Marta Gomez-Barrero\thanks{ORCID: \href{https://orcid.org/0000-0003-4581-5353}{0000-0003-4581-5353}}}
\affil[1]{BioML, Research Institute CODE, University of the Bundeswehr Munich, Munich, Germany}
\affil[2]{Amsterdam UMC, University of Amsterdam, Amsterdam, Netherlands}
\date{}
\begin{document}
\maketitle

\begin{abstract}
Electrocardiography (ECG) contains subject-specific morphology that can support biometric recognition, yet its use in image-based systems depends strongly on how the waveform is represented. We introduce representative-morphology heatmaps, a deterministic ECG-to-image representation adapted from the representative segment construction of ECGXtractor. Within each block of ten aligned beats, the five beats closest to the block mean are averaged into a fixed $400\times L$ matrix, which is then rendered either as a conventional trace or as a dense cardiac-time-by-lead heatmap. Since both images contain the same physiological samples, their comparison isolates the effect of representation.

The evaluation covers verification and closed-set identification on PTB, ECG-ID, and MIMIC-IV-ECG-DEMO. Five compact image models, including ZACH-ViT, are trained from scratch, while six ImageNet-pretrained CNN and transformer backbones examine model scale and visual transfer. Matched scratch and pretrained runs assess the effect of initialization, and four signal-domain methods provide direct references.

Heatmaps improve both FNMR operating points and both identification ranks in all 15 compact model and dataset comparisons, while mean equal error rate (EER) improves in 14 comparisons. Across the complete matched experiment, EER decreases by an average of 9.59 percentage points and Rank-1 increases by 24.69 points. ConvNeXt-Tiny reaches 2.43\% EER on PTB and 5.79\% on ECG-ID, whereas DeiT-Base reaches 14.92\% on MIMIC-DEMO. ImageNet initialization provides clear gains on the two multilead datasets but has a mixed effect on ECG-ID, and performance does not increase monotonically with model size. In particular, ZACH-ViT remains competitive with substantially larger models trained from scratch. The best heatmap systems approach the strongest signal EER on PTB and ECG-ID, while DeiT-Base is best on four metrics and tied for best on Rank-5 on MIMIC-DEMO. Lead-channel ablation further shows that useful channel combinations depend on the cohort and biometric task. Overall, representative-morphology heatmaps provide an effective image representation for ECG verification and identification.
\end{abstract}


\section{Introduction}
ECG biometrics exploit cardiac electrical activity as a physiological identity signal. Since ECG is naturally acquired by devices that maintain skin contact, it is relevant to wearable authentication, clinical terminals, continuous identity assurance, and related applications \cite{melzi2023ecgxtractor}. In these settings, recognition accuracy is only one part of the system design, which must also account for sensing requirements, signal representation, and model complexity.

Direct signal processing provides the natural reference for ECG, while signal-to-image encoding offers a complementary route that can use established vision architectures, pretrained models, and image-based deployment tools. However, the usefulness of this route depends on the image construction. Conventional ECG traces are intended for human inspection and become sparse after rasterization, because most pixels represent background rather than measured samples. A dense image can instead assign the pixel grid directly to the signal values, although a controlled comparison requires the underlying ECG morphology to remain unchanged.

Our pipeline begins with a representative morphology adapted from ECGXtractor \cite{melzi2023ecgxtractor}. R-peak-aligned beats are divided into blocks of ten, from which the five beats closest to the block mean are selected and averaged. This process produces a fixed matrix $S\in\R^{400\times L}$, where $L$ denotes the number of acquired leads. The same matrix is rendered as a stacked trace and as a dense heatmap whose axes encode cardiac-relative time and lead identity, while colour represents normalized amplitude. The transformation therefore reorganizes the ECG summary without changing its physiological content.

Previous studies have considered ECG colour maps, waveform images, convolutional networks, and vision transformers, but the present work addresses a different and more controlled question. We transform an ECGXtractor-style representative morphology into a dense time-by-lead image and generate the conventional trace from the identical numerical source. Filtering, R-peak localization, beat alignment, beat selection, and physiological samples are consequently shared between the two representations, allowing their effect on biometric recognition to be measured directly. The resulting evaluation also brings together compact models, visual pretraining, signal-domain references, verification, and identification within a common protocol.

Experiments are conducted on PTB, ECG-ID, and MIMIC-IV-ECG-DEMO \cite{ptb,ecgid,lugovaya2005,mimic,physionet}. Five compact architectures are trained from scratch on both traces and heatmaps to quantify the representation effect, after which six ImageNet-1K-pretrained backbones, ranging from 4.01M to 85.80M extractor parameters, are fine-tuned on the heatmaps. ConvNeXt-Tiny and DeiT-Base are additionally trained from scratch under the same 23-epoch protocol, thereby isolating the contribution of initialization. Direct signal baselines, exhaustive lead-channel ablation, and image degradation experiments complete the study.

The main contributions of this work are:
\begin{enumerate}[leftmargin=1.4em]
\item a deterministic heatmap construction based on representative ECG morphology, together with matched traces generated from the same numerical samples;
\item a harmonized biometric evaluation on three subject-disjoint datasets, covering both verification and closed-set identification;
\item a controlled trace-versus-heatmap experiment with five compact models, in which heatmaps improve both FNMR measures and both identification ranks in all 15 model-dataset comparisons;
\item an ImageNet-pretrained benchmark with six backbones, complemented by matched ConvNeXt-Tiny and DeiT-Base experiments that isolate the effect of initialization; and
\item a direct comparison with four signal-domain methods, supported by exhaustive lead-channel ablation and image degradation analyses.
\end{enumerate}

\section{Related Work and Positioning}
\subsection{ECG biometric recognition}
ECG biometric recognition has been investigated through fiducial descriptors, morphology templates, cross-correlation, metric learning, and deep neural networks. Melzi et al. review this literature and introduce ECGXtractor, which selects the five beats closest to the mean of ten aligned beats before averaging them into a representative 400-sample morphology \cite{melzi2023ecgxtractor}. ECGXtractor processes this summary numerically for verification and identification, whereas our work develops and evaluates its image-domain representation.

The influence of the acquired leads has also been established. Krasteva et al. showed that individual 12-lead projections differ in biometric reliability and that multilead combinations can improve verification \cite{krasteva2017}. At the model level, single-heartbeat CNNs demonstrate that compact one-dimensional networks can extract identity information directly from ECG \cite{alduwaile2021singleheartbeat}, while InceptionTime provides a strong general convolutional architecture for time-series classification \cite{fawaz2020}. These findings motivate our comparison with direct signal processing and our analysis of how lead-channel utility changes across cohorts and biometric tasks.

\subsection{ECG-to-image representations}
Image-based ECG biometrics predate the present work. Silva et al. arranged aligned heartbeat waveforms as rows of a colour map, while ELEKTRA constructs an electrocardiomatrix before CNN-based identification \cite{silva2013ecg,fusterbarcelo2022elektra}. In both approaches, the second image dimension indexes separate beats; by contrast, our heatmap uses physical ECG leads after the beats have been aggregated. Other studies rely on coupling images, wavelets, recurrence plots, Gramian angular fields, state-space representations, and time-frequency transformations, each of which preserves a different aspect of the original signal \cite{kim2020coupling,elboujnouni2022wavelet,ciocoiu2020spatial}.

Visual transfer learning has also been explored for ECG biometrics. AlDuwaile and Islam evaluated custom and pretrained CNNs on continuous-wavelet images \cite{alduwaile2021singleheartbeat}, whereas D'Angelis et al. fine-tuned an ImageNet-pretrained transformer on averaged single-lead waveform images for verification and identification \cite{dangelis2023ecgvit}. CardioIdNet provides a more recent compact image-based authentication model trained from scratch \cite{rossi2026cardioidnet}. The present contribution differs from these studies through the construction of the representative morphology and, crucially, the matched rendering design used to isolate the image representation.

Related representations have also been proposed outside the biometric domain. Bortolan maps raw multilead intervals to time-by-lead images for cardiac abnormality classification, while HeartBEiT pretrains a masked-image transformer on ECG images for diagnosis \cite{bortolan2023ecgdisplay,vaid2023heartbeit}. Together, these studies demonstrate the broader potential of ECG-derived images, although neither examines the representative-morphology transformation considered here.

Our method first aggregates morphologically central beats into one ECGXtractor-style summary and then assigns acquired-lead identity to the second image axis. Since the conventional trace and dense heatmap are generated from the same matrix, filtering, R-peak localization, beat alignment, beat selection, and physiological samples remain fixed. The comparison can therefore be attributed to the rendering rather than to differences in ECG preprocessing.

A trace remains sparse after rasterization because it was designed for human interpretation, and its appearance is affected by line width, antialiasing, spacing, and resolution. The heatmap instead maps the numerical matrix directly to image coordinates and uses colour to encode normalized amplitude. Adjacent rows preserve the lead order without implying an anatomical distance between leads. ECG-ID offers a particularly useful control in this respect: its heatmap contains only one acquired lead, so any improvement over the trace reflects dense rasterization without inter-lead structure.

\subsection{Compact models, model scale, and visual transfer}
The compact branch spans several inductive biases. ABMIL learns attention-weighted aggregation over image instances \cite{ilse2018}, TransMIL models relations between correlated instances \cite{shao2021}, and Compact Transformers reduce the data and parameter requirements of vision transformers \cite{hassani2021}. ZACH-ViT is a compact position-free transformer developed for low-data medical imaging, with a design that also supports CPU training \cite{angelakis2026,angelakis2026extending}. More recently, Patch-ABMIL, Compact-TransMIL, and ZACH-ViT were evaluated for iris presentation attack detection under unknown presentation attack instruments and controlled Gaussian noise, blur, and JPEG degradation \cite{angelakis2026irispad}. That study addresses attack detection rather than identity recognition, but provides a relevant cross-modality assessment of the same compact model family under biometric shift and image corruption. Together, these architectures test whether the proposed heatmap remains effective when the reference classifier contains fewer than 0.6M parameters.

The pretrained branch extends the investigation to established CNNs and vision transformers at substantially larger scales. EfficientNet and ResNet represent conventional convolutional families \cite{tan2019efficientnet,he2016resnet}, DeiT provides transformers at three model sizes \cite{touvron2021deit}, and ConvNeXt contributes a modern convolutional architecture \cite{liu2022convnext}. ImageNet-1K initialization is used to examine whether generic visual features transfer to ECG heatmaps \cite{deng2009imagenet}, while matched ConvNeXt-Tiny and DeiT-Base controls isolate initialization under a fixed training budget. The compact and large branches therefore describe complementary regimes of model footprint, architecture, and visual transfer.

\section{Materials and Methods}
\subsection{Datasets and harmonized cohorts}
Table~\ref{tab:datasets} summarizes the harmonized cohorts. Identity partitions are fixed and subject-disjoint, with learning restricted to the training identities and biometric evaluation performed on unseen test identities. Every model is trained for a fixed number of epochs, and neither the validation nor test partition is used for checkpoint or hyperparameter selection.

\begin{table*}[t]
\centering
\caption{Harmonized dataset and biometric protocol overview. ``Train/val/test'' reports subject counts. Mated/non-mated are the sampled verification comparisons under the fixed protocol.}
\label{tab:datasets}
\small
\resizebox{\textwidth}{!}{%
\begin{tabular}{lcccccccc}
\toprule
Dataset & Leads & Subjects & Records & Summary seg. & Train/val/test & Gallery & Probe & Mated/non-mated \\
\midrule
PTB & 12 & 290 & 549 & 7064 & 203/29/58 & 58 & 1462 & 174/870 \\
ECG-ID & 1 (Lead I) & 90 & 310 & 599 & 62/9/19 & 19 & 72 & 55/285 \\
MIMIC-DEMO & 12 & 77 & 486 & 510 & 55/9/13 & 13 & 107 & 35/195 \\
\bottomrule
\end{tabular}%
}
\end{table*}

Harmonization retains identities with at least two usable summary segments and common availability across the signal, trace, and heatmap representations, without changing the ECG-to-image transformation. No identity appears in more than one partition, every representation contains unique sample keys, and the trace and heatmap manifests correspond one-to-one by sample key. These controls exclude both identity and sample leakage.

\subsection{ECG preprocessing and summary-segment construction}
All ECGs are processed at 500 Hz, with recordings acquired at another frequency resampled accordingly. NeuroKit2 performs ECG cleaning and R-peak detection \cite{neurokit} using one synchronized reference channel: Lead I for PTB, Lead II for MIMIC-DEMO, and the available Lead I for ECG-ID. In each multilead recording, the resulting peak indices are applied to all channels simultaneously, and the first and last peaks are removed whenever at least three peaks are available.

Around each retained R-peak $r$, a 400-sample window is extracted,
\begin{equation}
B_r = X[r-160:r+240],
\end{equation}
which corresponds to 0.32 s before and 0.48 s after the R-peak. Windows that cross a recording boundary are discarded. PTB and MIMIC-DEMO beats are independently z-score normalized per lead, whereas ECG-ID follows the prepared single-lead pipeline in which the complete Lead-I recording is normalized before R-peak detection and beat extraction.

Valid heartbeats are partitioned into sequential, non-overlapping blocks of ten, and any remainder smaller than ten is discarded. For a block $\mathcal{B}=\{B_1,\ldots,B_{10}\}$, we calculate the element-wise mean
\begin{equation}
\bar B=\frac{1}{10}\sum_{j=1}^{10}B_j,
\end{equation}
The beats are then ranked by Euclidean distance after flattening, equivalently by the Frobenius norm $\|B_j-\bar B\|_F$. The five closest beats form $\mathcal{C}$ and are averaged:
\begin{equation}
S=\frac{1}{5}\sum_{B_j\in\mathcal{C}}B_j, \qquad S\in\R^{400\times L}.
\end{equation}
We refer to $S$ as a \emph{summary segment}, since it represents five morphologically central beats selected from ten consecutive valid beats. The same matrix is used for the trace, heatmap, summary-cosine baseline, 1D-CNN, and InceptionTime, while the non-overlapping block construction ensures that no beat contributes to more than one summary segment. The released ECGXtractor system is evaluated separately with its official preprocessing and pretrained model assets.

\subsection{Deterministic ECG-to-image transformation}
\begin{figure*}[t]
\centering
\includegraphics[width=0.98\textwidth]{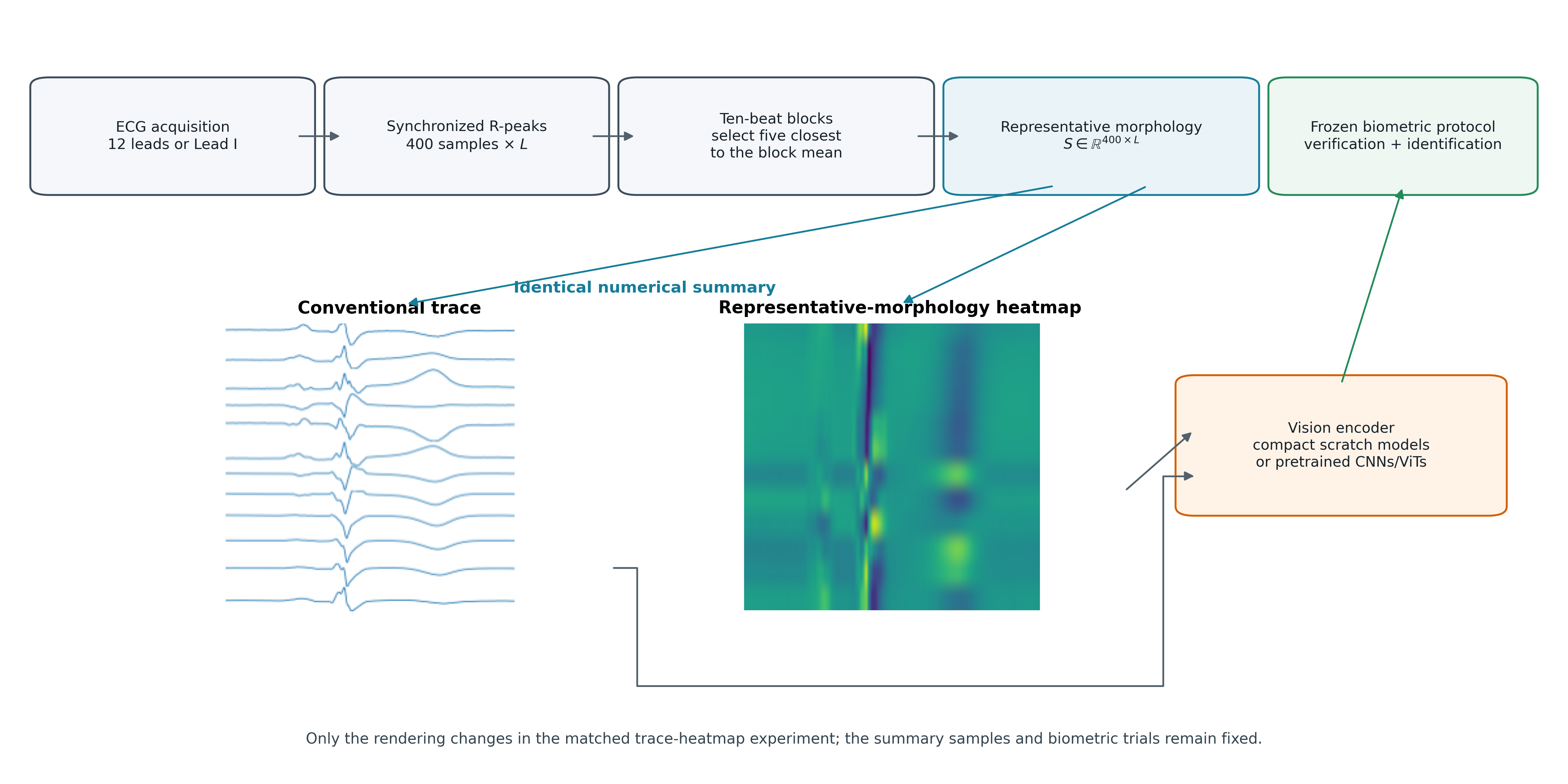}
\caption{Deterministic ECG-to-image biometric pipeline. A synchronized reference lead localizes R-peaks; all available leads are cut at the same indices. Ten consecutive valid beats are summarized by selecting the five closest to the block mean. The identical $400\times L$ representative morphology is rendered either as a sparse trace or as a dense time-lead heatmap before compact scratch-trained or pretrained vision encoding.}
\label{fig:pipeline}
\end{figure*}

Figure~\ref{fig:pipeline} illustrates the deterministic transformation. To construct a trace image, each column of $S$ is plotted as a waveform, with multilead inputs arranged in a vertical stack and ECG-ID represented by a single waveform. The renderer uses a line width of 0.6, removes axes, labels, and plotting padding, and produces a nominal $2.24\times2.24$ inch image at 100 dpi. The output is converted to RGB and resized only when necessary to obtain exactly $224\times224$ pixels.

For the heatmap, $S^\top\in\R^{L\times400}$ is rendered directly, with time on the horizontal axis and acquired leads on the vertical axis. We use `aspect=auto' and `origin=lower', retaining the standard order I, II, III, aVR, aVL, aVF, and V1-V6 for the 12-lead inputs. Viridis encodes normalized amplitude, with the colour range determined by the minimum and maximum of each summary matrix. Axes and padding are removed before RGB conversion, and the final image has $224\times224$ pixels. In ECG-ID, the renderer expands the single $1\times400$ row vertically without introducing any additional physiological information.

\subsection{Compact image models}
The compact roster was fixed before the final runs and spans attention MIL, patch attention, transformer reasoning over patches, compact convolutional tokenization, and ZACH-ViT. Compactness is defined by the parameter count of the reference classifier rather than by the dataset-specific identity head, and all five configurations remain below 0.6M parameters (Table~\ref{tab:models}).

\begin{table*}[t]
\centering
\caption{Compact image models. Parameter counts are the reference classifier counts used to define the sub-0.6M inclusion criterion; dataset-specific identity heads do not alter that inclusion rule.}
\label{tab:models}
\small
\begin{tabularx}{\textwidth}{lrrX}
\toprule
Model & Ref. params & Millions & Implementation \\
\midrule
ABMIL & 490,828 & 0.491 & Official ungated ABMIL; 64 grayscale $28\times28$ instances; attention pooling \\
Patch-ABMIL & 90,000 & 0.090 & 256 RGB $14\times14$ patches; 128-D patch features; attention pooling \\
Compact-TransMIL & 260,000 & 0.260 & 256 RGB $14\times14$ patches; 2 transformer blocks, 4 heads; GAP \\
CCT-2/3x2 & 284,000 & 0.284 & Official CCT-2/3$\times$2 configuration; $32\times32$ input \\
ZACH-ViT & 250,000 & 0.250 & 196 RGB $16\times16$ patches; no CLS/position token; adaptive residual projection; GAP \\
\bottomrule
\end{tabularx}
\end{table*}

ABMIL uses the official ungated attention architecture. Each $224\times224$ image is converted to grayscale and divided into an $8\times8$ bag of non-overlapping $28\times28$ instances, after which the attention-pooled representation $Z$ forms the biometric embedding. Patch-ABMIL retains the RGB image, extracts 256 non-overlapping $14\times14$ patches, projects them to 128 dimensions, and applies tanh attention pooling. Compact-TransMIL uses the same patches with two transformer blocks, four attention heads, and global average pooling, while CCT-2/3$\times$2 follows its published $32\times32$ configuration \cite{hassani2021}. ZACH-ViT processes 196 non-overlapping $16\times16$ patches without a class token or positional embedding; adaptive residual projections accommodate changes in representation width, and global average pooling produces the final embedding \cite{angelakis2026,angelakis2026extending}.

All compact image models are trained from scratch on CPU for 97 epochs using Adam, a learning rate of $10^{-4}$, sparse categorical cross-entropy, and the seeds $\{3,5,7,11,13\}$. The batch size is 16, except for the official ABMIL adapter, which uses at most four image bags because every image contains 64 instances. The seed controls initialization and training order, while the dataset partitions and biometric comparisons remain fixed. After training, the identity classifier is discarded and the designated internal representation is used for biometric scoring.

\subsection{ImageNet-pretrained heatmap scale benchmark}
While the compact benchmark focuses on small CPU-executable models, a second benchmark evaluates larger architectures with visual pretraining. It includes EfficientNet-B0, DeiT-Tiny, ResNet-18, DeiT-Small, ConvNeXt-Tiny, and DeiT-Base, which together span three CNNs, three transformers, and 4.01M to 85.80M extractor parameters \cite{tan2019efficientnet,touvron2021deit,he2016resnet,liu2022convnext}.

\begin{table*}[t]
\centering
\caption{ImageNet-pretrained heatmap benchmark. Parameter counts exclude the dataset-specific identity-classification head and refer to the retained feature extractor. ``Full FT'' denotes full-network fine-tuning; no backbone is frozen.}
\label{tab:pretrained_models}
\small
\resizebox{\textwidth}{!}{%
\begin{tabular}{lllrrl}
\toprule
Model & Family & \texttt{timm} identifier & Extractor params (M) & Embedding dim. & Protocol \\
\midrule
EfficientNet-B0 & CNN & \texttt{efficientnet\_b0.ra\_in1k} & 4.008 & 1280 & Full FT, 23 epochs \\
DeiT-Tiny & ViT & \texttt{deit\_tiny\_patch16\_224.fb\_in1k} & 5.524 & 192 & Full FT, 23 epochs \\
ResNet-18 & CNN & \texttt{resnet18.a1\_in1k} & 11.177 & 512 & Full FT, 23 epochs \\
DeiT-Small & ViT & \texttt{deit\_small\_patch16\_224.fb\_in1k} & 21.666 & 384 & Full FT, 23 epochs \\
ConvNeXt-Tiny & CNN & \texttt{convnext\_tiny.fb\_in1k} & 27.820 & 768 & Full FT, 23 epochs \\
DeiT-Base & ViT & \texttt{deit\_base\_patch16\_224.fb\_in1k} & 85.799 & 768 & Full FT, 23 epochs \\
\bottomrule
\end{tabular}%
}
\end{table*}

Each backbone uses its specified ImageNet-1K weights \cite{deng2009imagenet}, and a new identity-classification head is added for each dataset before the complete network is fine-tuned. Training uses Adam with a learning rate of $10^{-4}$, zero weight decay, batch size 16, identity-supervised cross-entropy, and the same five seeds. Every run lasts 23 epochs, without early stopping, learning-rate scheduling, or validation-based checkpoint selection.

Heatmaps are converted to RGB, resized to $224\times224$, and normalized with the ImageNet channel statistics. During training, brightness and contrast jitter of 0.05 are applied, whereas test images receive only resizing and normalization. After epoch 23, the classifier is discarded and the pre-logit representation forms the biometric embedding. L2-normalized embeddings are compared by cosine similarity under the fixed protocol; ECGXtractor verification is the only exception and uses the released Siamese probability. The larger models use GPU support, so the CPU claim applies exclusively to the compact image branch.

\subsection{Matched initialization ablation}
ConvNeXt-Tiny and DeiT-Base represent the CNN and transformer branches in the initialization ablation. Each pretrained result is paired with a scratch run from the same \texttt{timm} architecture, with the seed, epoch count, and protocol manifest matched before comparison.

Initialization is the only factor that changes within each pair: the heatmaps, subject partitions, identity heads, transforms, optimizer, learning rate, batch size, and 23-epoch budget remain identical, as do embedding extraction and biometric evaluation. Each architecture therefore contributes 15 matched pretrained-scratch pairs. Since the compact models follow a separate 97-epoch CPU protocol, comparisons across the two training schedules describe achieved accuracy relative to model size rather than a controlled scaling experiment.

\subsection{Signal-domain comparators}
Four signal-domain approaches use the same harmonized summary segments.
\paragraph{Summary-segment cosine.} The $400\times L$ summary matrix is flattened and L2-normalized before cosine similarity is calculated directly. This non-learned baseline also supplies the representation used in the exhaustive lead-subset analysis.
\paragraph{ECGXtractor.} We evaluate the released pretrained feature extractor and Siamese verifier without retraining \cite{melzi2023ecgxtractor,ecgxtractor}. PTB and MIMIC use the official 12-lead model, whereas ECG-ID uses the official Lead-I configuration and latent slice. Verification relies on the Siamese matching probability after the saved preprocessing, and identification uses cosine similarity between the extracted features.
\paragraph{1D-CNN.} A compact two-layer Conv1D network is applied to the common $400\times L$ summary \cite{alduwaile2021singleheartbeat}. The layers contain 16 and 32 filters with kernel size 3 and are each followed by max pooling; a 100-D dense embedding, dropout of 0.2, and an identity head complete the model.
\paragraph{InceptionTime.} We use a single InceptionTime-style network rather than an ensemble \cite{fawaz2020}. It contains six inception modules, a bottleneck size of 32, and residual shortcuts after every three modules, with the global average pooled representation serving as the biometric embedding.

The two learned signal models use the same five seeds and are trained for 97 epochs with Adam at $10^{-4}$, batch size 16, and identity-supervised cross-entropy. GPU execution is permitted for this branch, since the CPU claim is limited to the compact image models.

\subsection{Biometric protocols and metrics}
We follow ISO biometric terminology \cite{iso19795}, defining a \emph{mated} comparison between samples of the same identity and a \emph{non-mated} comparison between different identities. For each test subject, the fixed protocol samples up to three mated and 15 non-mated comparisons using seed 3. Mated comparisons always contain distinct summary segments, non-mated comparisons always contain different identities, and the protocol does not require acquisition from separate recording sessions.

Embedding-based systems use cosine similarity. False match rate (FMR) and false non-match rate (FNMR) are computed with a fixed detection error trade-off implementation, and EER is reported as the mean of both error rates at the threshold that minimizes $|\mathrm{FMR}-\mathrm{FNMR}|$. We additionally report FNMR at target FMR values of 10\% and 1\%, selecting at each target the threshold with the lowest FNMR among those that satisfy the prescribed FMR limit.

Closed-set identification uses one gallery sample for each test identity, selected after seed-3 shuffling, while every remaining row becomes a probe. Gallery and probe samples are therefore disjoint. Probe embeddings are ranked by cosine similarity, and Rank-1 and Rank-5 indicate whether the correct identity occurs within the first one or five positions; the complete cumulative match characteristic is also retained.

\subsection{Lead-channel scoring ablation}
PTB and MIMIC-DEMO follow the standard channel order I, II, III, aVR, aVL, aVF, and V1-V6. We first evaluate each channel and the complete 12-channel summary under the cosine protocol, followed by all $2^{12}-1=4095$ non-empty subsets in each dataset. The implementation caches channel-specific dot products and norms, was validated against the exact flattened embeddings, and recomputes every reported subset with the exact scoring functions.

Channels are removed only after the 12-channel summaries have been created, which means that representative beats are selected using the full multilead Frobenius distance. The experiment is consequently a scoring-channel ablation rather than a simulation of prospective reduced-electrode acquisition. R-peaks are also localized from the synchronized reference channel defined above, so a label such as ``V1'' indicates that V1 is retained for cosine scoring, not that the complete preprocessing pipeline uses only that sensor. Since ECG channels and physical electrodes are not equivalent, all results are expressed in terms of channel counts.

Cross-dataset concordance is measured by ranking identical lead combinations according to EER and Rank-1 for every subset size $k$, followed by Spearman correlation between the two datasets. Because the exhaustive subsets overlap and are not independent observations, these correlations are used descriptively to assess portability across cohorts.

\subsection{Heatmap degradation stress test}
ZACH-ViT and Compact-TransMIL are evaluated without retraining on clean and corrupted heatmaps. The experiment considers Gaussian noise with $\sigma=0.03$ and 0.06, Gaussian blur with radius 1 and 2, JPEG compression at quality 70 and 40, and resolution reduction to $112\times112$ and $64\times64$ followed by restoration to $224\times224$.

Pillow performs RGB blur and JPEG encoding and decoding, while bilinear interpolation is used for the resolution changes. NumPy adds noise in the $[0,1]$ pixel domain, clips the result, and initializes its generator with the corresponding training seed. We load the final 97-epoch weights, verify every model against the clean benchmark, and then run inference on CPU for all corruptions. The eight degraded conditions are averaged within each seed before the mean and standard deviation are reported over the five seeds.

\section{Results}

\subsection{Dense heatmaps improve matched trace results}
\begin{figure*}[t]
\centering
\includegraphics[width=0.94\textwidth]{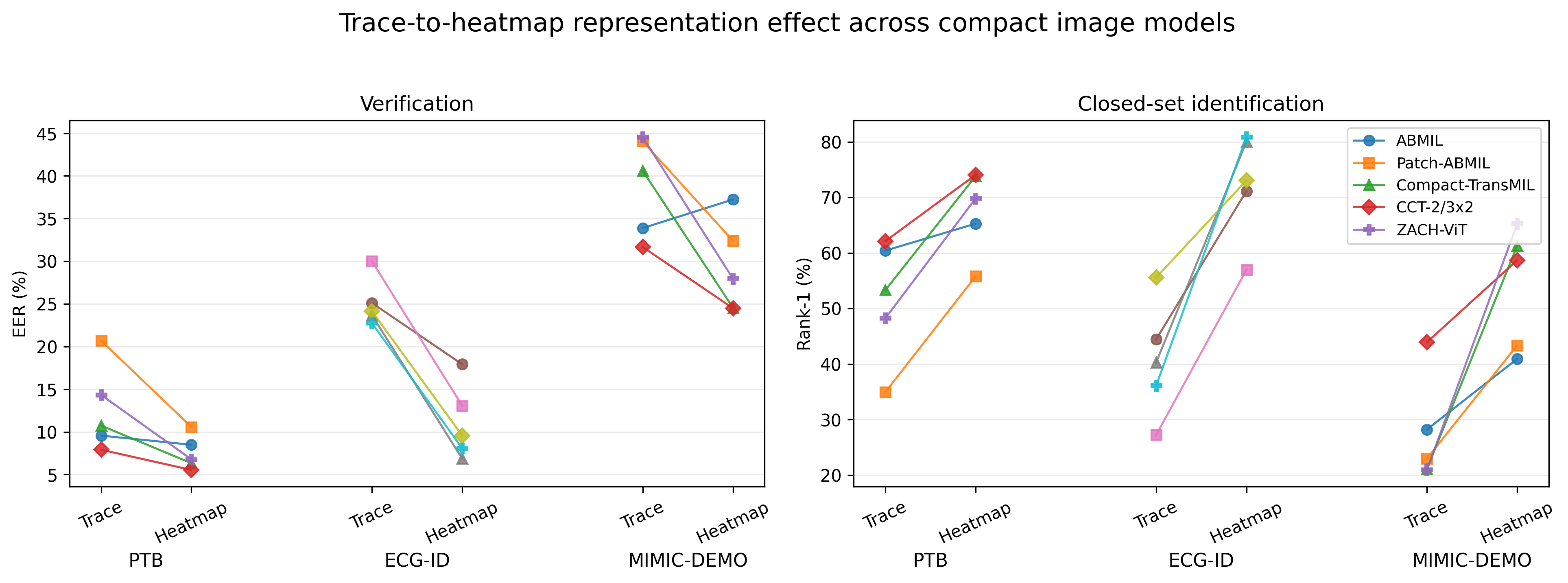}
\caption{Paired representation effect for the five compact architectures across the three datasets. Each line connects the five-seed mean trace and heatmap result for the same architecture and fixed training protocol. Lower EER and higher Rank-1 are better. Heatmaps improve all 15 mean identification cells and 14 of 15 mean EER cells.}
\label{fig:rep}
\end{figure*}

Table~\ref{tab:representation} and Figure~\ref{fig:rep} compare traces and heatmaps generated from the same ECG summaries, with the model family, training budget, subject partition, and biometric protocol held fixed. Heatmaps reduce \fnmrten{} and \fnmrone{} in all 15 model-dataset comparisons, while also increasing Rank-1 and Rank-5 in every case. Mean EER improves in 14 of 15 comparisons. When averaged over the complete experiment, EER decreases by 9.59 percentage points and Rank-1 increases by 24.69 points; the corresponding average changes are 23.25 points for \fnmrten, 24.66 points for \fnmrone, and 17.97 points for Rank-5.

The representation effect is observed in all three datasets, with mean EER reductions of 5.10 points on PTB, 14.04 points on ECG-ID, and 9.62 points on MIMIC-DEMO. The respective Rank-1 increases are 15.94, 31.67, and 26.47 points. MIMIC-DEMO ABMIL provides the only exception in EER, changing from $33.88\pm6.19$\% for the trace to $37.24\pm3.14$\% for the heatmap, although both FNMR measures and both identification ranks still improve.

ECG-ID provides the clearest single-lead control because its heatmap contains no inter-lead structure. ZACH-ViT improves from $22.75\pm3.66$\% to $8.07\pm0.96$\% EER, while Rank-1 increases from $36.11\pm6.05$\% to $80.83\pm1.16$\%. Compact-TransMIL shows the same pattern, reducing EER from $23.68\pm4.10$\% to $6.86\pm2.06$\% and increasing Rank-1 from $40.28\pm11.74$\% to $80.00\pm2.52$\%. These results indicate that dense rasterization improves access to the same normalized waveform even when no information is available across leads.

\begin{landscape}
\begin{table}[p]
\centering
\caption{Compact image benchmark over five seeds. Trace and heatmap models use the same ECG summaries and 97-epoch scratch protocol. Values are mean $\pm$ sample SD. Complete operating points appear in Appendix~\ref{app:representation-full}.}
\label{tab:representation}
\scriptsize
\begin{tabular}{llcccccc}
\toprule
Dataset & Model & Trace EER & Heatmap EER & Heatmap FNMR@$\leq$10 & Heatmap FNMR@$\leq$1 & Trace R1 & Heatmap R1 \\
\midrule
PTB & ABMIL & 9.56 $\pm$ 2.42 & 8.49 $\pm$ 1.78 & 7.93 $\pm$ 2.38 & 18.05 $\pm$ 2.71 & 60.41 $\pm$ 4.66 & 65.27 $\pm$ 4.05 \\
PTB & Patch-ABMIL & 20.69 $\pm$ 2.77 & 10.57 $\pm$ 0.87 & 11.03 $\pm$ 1.43 & 25.40 $\pm$ 2.77 & 34.95 $\pm$ 9.80 & 55.77 $\pm$ 1.72 \\
PTB & Compact-TransMIL & 10.74 $\pm$ 1.13 & 6.33 $\pm$ 0.91 & 5.06 $\pm$ 1.18 & 13.56 $\pm$ 4.06 & 53.30 $\pm$ 5.93 & 73.82 $\pm$ 4.77 \\
PTB & CCT-2/3x2 & 7.91 $\pm$ 2.63 & \textbf{5.54 $\pm$ 1.20} & 4.25 $\pm$ 1.19 & 12.41 $\pm$ 1.32 & 62.12 $\pm$ 11.86 & \textbf{74.05 $\pm$ 3.67} \\
PTB & ZACH-ViT & 14.33 $\pm$ 3.26 & 6.78 $\pm$ 0.49 & 5.40 $\pm$ 1.04 & 14.02 $\pm$ 1.04 & 48.24 $\pm$ 4.36 & 69.79 $\pm$ 2.17 \\
\midrule
ECG-ID & ABMIL & 25.11 $\pm$ 5.91 & 17.93 $\pm$ 1.37 & 25.09 $\pm$ 3.25 & 60.73 $\pm$ 11.54 & 44.44 $\pm$ 18.89 & 71.11 $\pm$ 6.16 \\
ECG-ID & Patch-ABMIL & 30.00 $\pm$ 3.82 & 13.07 $\pm$ 1.45 & 16.73 $\pm$ 3.25 & 49.45 $\pm$ 2.37 & 27.22 $\pm$ 2.88 & 56.94 $\pm$ 1.96 \\
ECG-ID & Compact-TransMIL & 23.68 $\pm$ 4.10 & \textbf{6.86 $\pm$ 2.06} & 5.09 $\pm$ 3.73 & 32.73 $\pm$ 11.13 & 40.28 $\pm$ 11.74 & 80.00 $\pm$ 2.52 \\
ECG-ID & CCT-2/3x2 & 24.14 $\pm$ 7.19 & 9.57 $\pm$ 2.25 & 9.09 $\pm$ 3.64 & 33.45 $\pm$ 9.15 & 55.56 $\pm$ 12.88 & 73.06 $\pm$ 5.25 \\
ECG-ID & ZACH-ViT & 22.75 $\pm$ 3.66 & 8.07 $\pm$ 0.96 & 4.00 $\pm$ 2.99 & 40.36 $\pm$ 9.48 & 36.11 $\pm$ 6.05 & \textbf{80.83 $\pm$ 1.16} \\
\midrule
MIMIC-DEMO & ABMIL & 33.88 $\pm$ 6.19 & 37.24 $\pm$ 3.14 & 61.14 $\pm$ 9.17 & 81.71 $\pm$ 7.45 & 28.22 $\pm$ 5.62 & 40.93 $\pm$ 6.39 \\
MIMIC-DEMO & Patch-ABMIL & 44.05 $\pm$ 3.32 & 32.39 $\pm$ 3.86 & 54.29 $\pm$ 4.52 & 85.71 $\pm$ 5.71 & 22.99 $\pm$ 6.79 & 43.36 $\pm$ 4.26 \\
MIMIC-DEMO & Compact-TransMIL & 40.59 $\pm$ 3.73 & 24.54 $\pm$ 2.47 & 40.57 $\pm$ 9.35 & 64.57 $\pm$ 7.98 & 21.12 $\pm$ 4.84 & 61.31 $\pm$ 4.56 \\
MIMIC-DEMO & CCT-2/3x2 & 31.66 $\pm$ 5.21 & \textbf{24.49 $\pm$ 4.64} & 44.00 $\pm$ 7.45 & 71.43 $\pm$ 10.88 & 43.93 $\pm$ 4.67 & 58.69 $\pm$ 9.49 \\
MIMIC-DEMO & ZACH-ViT & 44.54 $\pm$ 6.28 & 27.95 $\pm$ 4.74 & 38.29 $\pm$ 10.42 & 72.57 $\pm$ 5.57 & 20.93 $\pm$ 10.08 & \textbf{65.23 $\pm$ 3.40} \\
\bottomrule
\end{tabular}
\end{table}
\end{landscape}

The strongest compact architecture varies across datasets and metrics. CCT-2/3$\times$2 gives the lowest mean heatmap EER on PTB and MIMIC-DEMO, Compact-TransMIL gives the lowest mean EER on ECG-ID, and ZACH-ViT provides the highest compact-model Rank-1 on ECG-ID and MIMIC-DEMO. The consistent advantage of the heatmap is therefore not tied to a single architecture.

\subsection{Pretrained heatmap scale benchmark}
\begin{landscape}
\begin{figure}[p]
\centering
\includegraphics[width=0.92\linewidth]{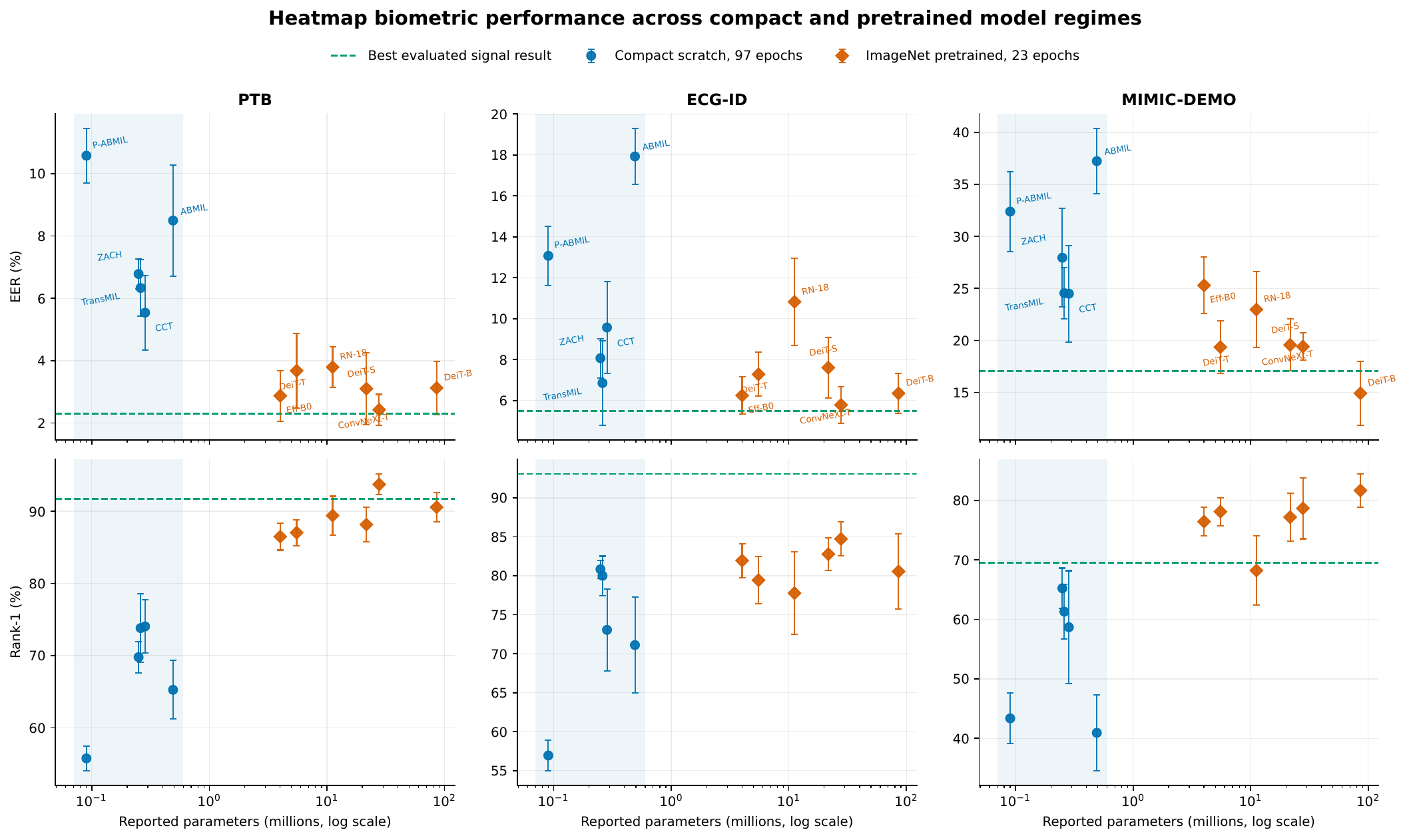}
\caption{Heatmap performance across compact scratch-trained and ImageNet-pretrained models. Error bars show sample SD over five seeds. Dashed lines indicate the strongest signal-domain value for each metric. Parameter-count conventions follow Table~\ref{tab:pretrained-scale}.}
\label{fig:pretrained-scale}
\end{figure}
\end{landscape}

The ImageNet-pretrained screen contains 90 successful runs. As shown in Table~\ref{tab:pretrained-scale} and Figure~\ref{fig:pretrained-scale}, the strongest architecture depends on the dataset: ConvNeXt-Tiny gives the lowest mean EER on PTB and ECG-ID at $2.43\pm0.50$\% and $5.79\pm0.89$\%, respectively, whereas DeiT-Base leads MIMIC-DEMO with $14.92\pm3.08$\%. Their corresponding Rank-1 values are $93.72\pm1.44$\%, $84.72\pm2.20$\%, and $81.68\pm2.77$\%.

Performance is not monotonically related to model size. ConvNeXt-Tiny outperforms the larger DeiT-Base on PTB and also provides the lowest mean EER on ECG-ID, while DeiT-Base becomes the strongest model on MIMIC-DEMO. Within the smaller transformers, DeiT-Tiny slightly outperforms DeiT-Small on both displayed metrics, and EfficientNet-B0 is competitive on PTB and ECG-ID but substantially weaker on MIMIC-DEMO. The complete five-metric results are reported in Appendix~\ref{app:pretrained-full}.

\begin{table*}[t]
\centering
\caption{Heatmap scale benchmark. Cells report EER / Rank-1 (\%, mean $\pm$ sample SD over five seeds). ZACH-ViT uses its reference-classifier count. The larger models use extractor counts without the identity head. ZACH-ViT uses a 97-epoch scratch protocol. The pretrained models use 23 epochs.}
\label{tab:pretrained-scale}
\scriptsize
\resizebox{\textwidth}{!}{%
\begin{tabular}{llrccc}
\toprule
Model & Initialization/epochs & Reported params (M) & PTB & ECG-ID & MIMIC-DEMO \\
\midrule
ZACH-ViT & scratch/97 & 0.25 &
$6.78{\pm}0.49/69.79{\pm}2.17$ &
$8.07{\pm}0.96/80.83{\pm}1.16$ &
$27.95{\pm}4.74/65.23{\pm}3.40$ \\
EfficientNet-B0 & ImageNet-1K/23 & 4.01 &
$2.87{\pm}0.81/86.48{\pm}1.87$ &
$6.25{\pm}0.90/81.94{\pm}2.20$ &
$25.29{\pm}2.72/76.45{\pm}2.42$ \\
DeiT-Tiny & ImageNet-1K/23 & 5.52 &
$3.68{\pm}1.20/87.03{\pm}1.81$ &
$7.29{\pm}1.08/79.44{\pm}3.01$ &
$19.36{\pm}2.55/78.13{\pm}2.35$ \\
ResNet-18 & ImageNet-1K/23 & 11.18 &
$3.79{\pm}0.66/89.40{\pm}2.69$ &
$10.82{\pm}2.12/77.78{\pm}5.29$ &
$22.97{\pm}3.67/68.22{\pm}5.84$ \\
DeiT-Small & ImageNet-1K/23 & 21.67 &
$3.10{\pm}1.16/88.15{\pm}2.41$ &
$7.61{\pm}1.48/82.78{\pm}2.11$ &
$19.56{\pm}2.50/77.20{\pm}4.00$ \\
ConvNeXt-Tiny & ImageNet-1K/23 & 27.82 &
$\mathbf{2.43}{\pm}0.50/\mathbf{93.72}{\pm}1.44$ &
$\mathbf{5.79}{\pm}0.89/\mathbf{84.72}{\pm}2.20$ &
$19.41{\pm}1.33/78.69{\pm}5.14$ \\
DeiT-Base & ImageNet-1K/23 & 85.80 &
$3.13{\pm}0.85/90.56{\pm}2.01$ &
$6.35{\pm}0.97/80.56{\pm}4.81$ &
$\mathbf{14.92}{\pm}3.08/\mathbf{81.68}{\pm}2.77$ \\
\bottomrule
\end{tabular}%
}
\end{table*}

Despite its compact design, ZACH-ViT remains competitive across the three datasets. Its 0.25M-parameter configuration is more than two orders of magnitude smaller than DeiT-Base, yet on ECG-ID it performs close to several pretrained backbones. This result places ZACH-ViT on a favourable accuracy and model-footprint trade-off.

\subsection{ImageNet initialization helps the multilead cohorts but is mixed for one lead}
\begin{landscape}
\begin{figure}[p]
\centering
\includegraphics[width=0.90\linewidth]{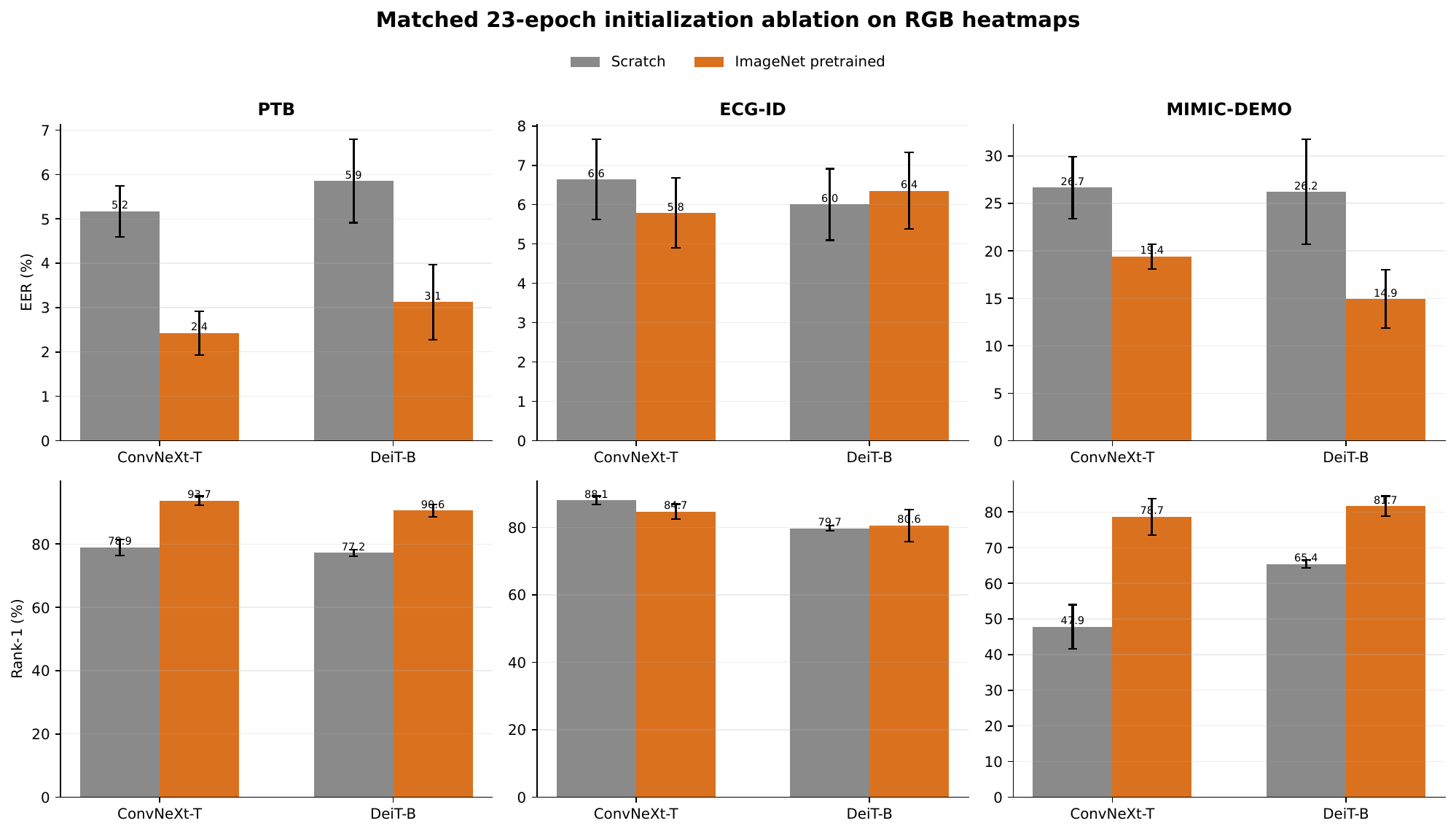}
\caption{Matched 23-epoch initialization ablation for ConvNeXt-Tiny and DeiT-Base. Bars show mean EER and Rank-1; error bars show sample SD over five seeds. Transfer is strongly beneficial on PTB and MIMIC-DEMO but limited or mixed on single-lead ECG-ID.}
\label{fig:initialization}
\end{figure}
\end{landscape}

The ConvNeXt-Tiny and DeiT-Base ablations isolate initialization under a common 23-epoch protocol, with the paired results reported in Table~\ref{tab:initialization} and Figure~\ref{fig:initialization}.

On PTB, pretraining reduces ConvNeXt-Tiny EER from $5.17\pm0.57$\% to $2.43\pm0.50$\% and increases Rank-1 from $78.92\pm2.53$\% to $93.72\pm1.44$\%. DeiT-Base follows the same trend, with EER decreasing from $5.85\pm0.94$\% to $3.13\pm0.85$\% and Rank-1 increasing from $77.21\pm1.12$\% to $90.56\pm2.01$\%. Both architectures improve in EER for every paired seed.

The effect is also pronounced on MIMIC-DEMO. ConvNeXt-Tiny reduces EER from $26.66\pm3.27$\% to $19.41\pm1.33$\% and increases Rank-1 from $47.85\pm6.14$\% to $78.69\pm5.14$\%, while DeiT-Base reduces EER from $26.22\pm5.51$\% to $14.92\pm3.08$\% and increases Rank-1 from $65.42\pm1.14$\% to $81.68\pm2.77$\%. As on PTB, EER improves for every paired seed in both models.

Single-lead ECG-ID follows a different pattern. Pretraining changes ConvNeXt-Tiny EER from $6.64\pm1.01$\% to $5.79\pm0.89$\%, but Rank-1 decreases from $88.06\pm1.24$\% to $84.72\pm2.20$\%. For DeiT-Base, EER changes from $6.00\pm0.91$\% to $6.35\pm0.97$\%, while Rank-1 changes from $79.72\pm0.76$\% to $80.56\pm4.81$\%. Visual pretraining is therefore consistently effective on the two multilead datasets, whereas its contribution on ECG-ID is limited and metric-dependent.

\begin{table*}[t]
\centering
\caption{Matched scratch-versus-ImageNet initialization ablation under the same 23-epoch heatmap protocol. Each cell reports scratch $\rightarrow$ ImageNet-1K performance (\%, mean $\pm$ sample SD over five seeds). Appendix~\ref{app:initialization-full} reports all operating points.}
\label{tab:initialization}
\small
\begin{tabular}{llcc}
\toprule
Model & Dataset & EER: scratch $\rightarrow$ pretrained & Rank-1: scratch $\rightarrow$ pretrained \\
\midrule
ConvNeXt-Tiny & PTB & $5.17\pm0.57 \rightarrow 2.43\pm0.50$ & $78.92\pm2.53 \rightarrow 93.72\pm1.44$ \\
ConvNeXt-Tiny & ECG-ID & $6.64\pm1.01 \rightarrow 5.79\pm0.89$ & $88.06\pm1.24 \rightarrow 84.72\pm2.20$ \\
ConvNeXt-Tiny & MIMIC-DEMO & $26.66\pm3.27 \rightarrow 19.41\pm1.33$ & $47.85\pm6.14 \rightarrow 78.69\pm5.14$ \\
\midrule
DeiT-Base & PTB & $5.85\pm0.94 \rightarrow 3.13\pm0.85$ & $77.21\pm1.12 \rightarrow 90.56\pm2.01$ \\
DeiT-Base & ECG-ID & $6.00\pm0.91 \rightarrow 6.35\pm0.97$ & $79.72\pm0.76 \rightarrow 80.56\pm4.81$ \\
DeiT-Base & MIMIC-DEMO & $26.22\pm5.51 \rightarrow 14.92\pm3.08$ & $65.42\pm1.14 \rightarrow 81.68\pm2.77$ \\
\bottomrule
\end{tabular}
\end{table*}

The compact models remain competitive with much larger scratch-trained architectures. On MIMIC-DEMO, CCT and Compact-TransMIL obtain 24.49\% and 24.54\% EER, compared with 26.66\% and 26.22\% for scratch ConvNeXt-Tiny and DeiT-Base. ZACH-ViT reaches 65.23\% Rank-1 in the same dataset, close to the 65.42\% of scratch DeiT-Base, and on ECG-ID it reaches 80.83\%, compared with 79.72\% for scratch DeiT-Base. Since the training schedules differ, these comparisons concern achieved accuracy and model footprint rather than controlled model scaling.

\subsection{Heatmaps are competitive with direct signal-domain methods}
Table~\ref{tab:signal} compares the four signal-domain references with the pretrained heatmap model that obtains the lowest mean EER in each dataset, using a common selection rule across the six-model screen.

On PTB, ECGXtractor gives the lowest EER at 2.30\%, followed closely by ConvNeXt-Tiny at $2.43\pm0.50$\%. The heatmap model nevertheless provides the lowest \fnmrten{} at $1.84\pm0.94$\% and the highest Rank-1 at $93.72\pm1.44$\%, whereas ECGXtractor remains stronger at the 1\% FMR target.

The 1D-CNN gives the lowest ECG-ID EER at $5.50\pm0.19$\%, with ConvNeXt-Tiny reaching $5.79\pm0.89$\%. ConvNeXt-Tiny is strongest at \fnmrten{} with $2.18\pm1.52$\%, although the 1D-CNN performs better at the 1\% FMR target and summary cosine gives the highest Rank-1 at 93.06\%.

MIMIC-DEMO produces the strongest image-domain result, as DeiT-Base is best on four measures and tied for best on Rank-5. Its EER is $14.92\pm3.08$\%, compared with 17.03\% for ECGXtractor, while \fnmrten{} is $20.00\pm5.35$\%, compared with 22.86\%, and \fnmrone{} is $54.86\pm14.76$\%, compared with 57.14\%. Rank-1 reaches $81.68\pm2.77$\%, exceeding the $69.53\pm5.67$\% of the 1D-CNN, and Rank-5 reaches $96.64\pm1.07$\%.

\begin{landscape}
\begin{table}[p]
\centering
\caption{Signal-domain references and the pretrained heatmap model with the lowest mean EER in each dataset. Learned methods report mean $\pm$ sample SD over five seeds. Summary cosine and pretrained ECGXtractor are single-run references. Lower EER and FNMR are better. Higher Rank-1 and Rank-5 are better. Bold marks the best value within each dataset.}
\label{tab:signal}
\scriptsize
\begin{tabular}{llccccc}
\toprule
Dataset & Method & EER (\%) & FNMR@$\leq$10 (\%) & FNMR@$\leq$1 (\%) & Rank-1 (\%) & Rank-5 (\%) \\
\midrule
PTB & Summary cosine & 3.39 & 2.30 & 6.32 & 90.15 & 96.72 \\
PTB & ECGXtractor & \textbf{2.30} & 2.30 & \textbf{2.30} & 91.72 & 97.67 \\
PTB & 1D-CNN & $4.15\pm0.48$ & $1.95\pm0.87$ & $6.78\pm0.94$ & $87.51\pm1.48$ & $96.09\pm1.17$ \\
PTB & InceptionTime & $5.75\pm0.70$ & $4.37\pm1.04$ & $10.69\pm0.66$ & $71.15\pm3.65$ & $89.22\pm1.99$ \\
PTB & ConvNeXt-Tiny heatmap & $2.43\pm0.50$ & \textbf{$1.84\pm0.94$} & $3.79\pm0.31$ & \textbf{$93.72\pm1.44$} & \textbf{$98.11\pm0.84$} \\
\midrule
ECG-ID & Summary cosine & 9.11 & 7.27 & 16.36 & \textbf{93.06} & \textbf{98.61} \\
ECG-ID & ECGXtractor & 7.32 & 3.64 & 27.27 & 88.89 & 94.44 \\
ECG-ID & 1D-CNN & \textbf{$5.50\pm0.19$} & $3.27\pm1.52$ & \textbf{$13.45\pm3.54$} & $88.33\pm1.58$ & $93.89\pm0.76$ \\
ECG-ID & InceptionTime & $5.53\pm1.70$ & $3.27\pm1.52$ & $24.73\pm8.87$ & $83.06\pm3.01$ & $94.44\pm0.00$ \\
ECG-ID & ConvNeXt-Tiny heatmap & $5.79\pm0.89$ & \textbf{$2.18\pm1.52$} & $25.82\pm4.71$ & $84.72\pm2.20$ & $95.56\pm2.28$ \\
\midrule
MIMIC-DEMO & Summary cosine & 28.90 & 37.14 & 68.57 & 61.68 & 80.37 \\
MIMIC-DEMO & ECGXtractor & 17.03 & 22.86 & 57.14 & 61.68 & 80.37 \\
MIMIC-DEMO & 1D-CNN & $21.73\pm2.99$ & $33.71\pm3.13$ & $69.14\pm7.40$ & $69.53\pm5.67$ & $96.07\pm1.54$ \\
MIMIC-DEMO & InceptionTime & $18.92\pm2.73$ & $29.71\pm8.48$ & $70.86\pm7.67$ & $61.12\pm4.36$ & $91.40\pm5.18$ \\
MIMIC-DEMO & DeiT-Base heatmap & \textbf{$14.92\pm3.08$} & \textbf{$20.00\pm5.35$} & \textbf{$54.86\pm14.76$} & \textbf{$81.68\pm2.77$} & \textbf{$96.64\pm1.07$} \\
\bottomrule
\end{tabular}
\end{table}
\end{landscape}

\begin{figure*}[t]
\centering
\includegraphics[width=0.96\textwidth]{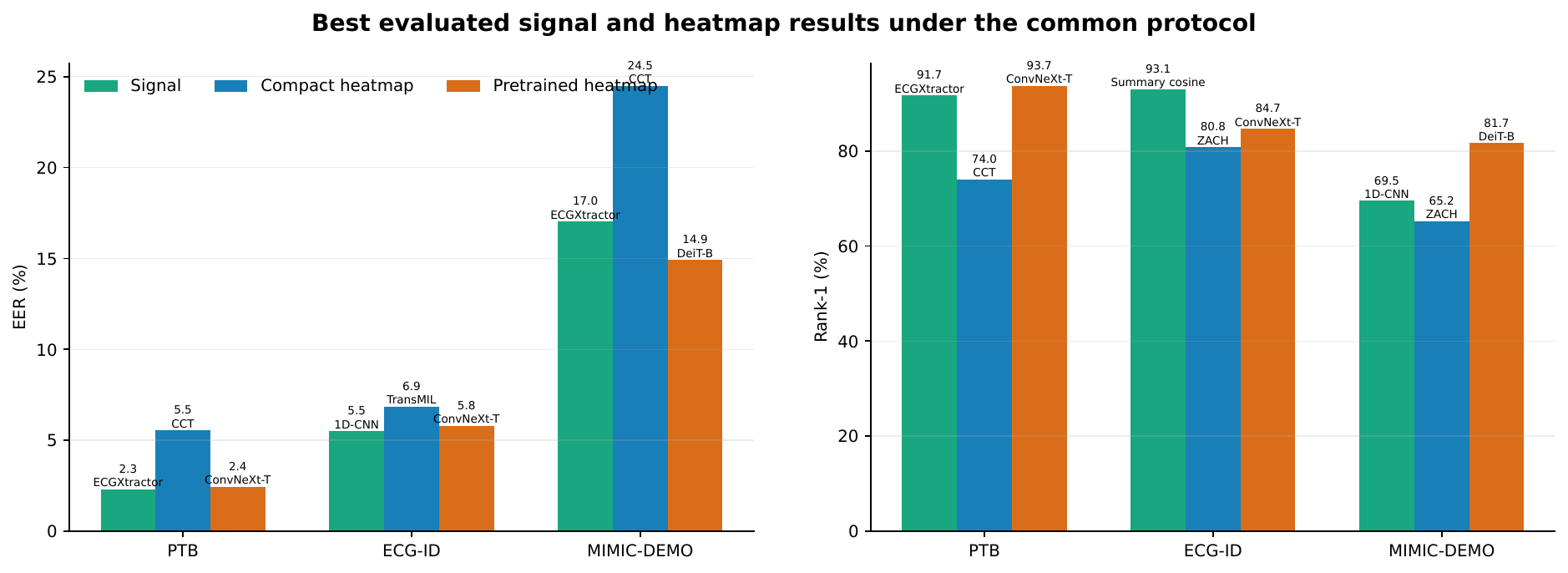}
\caption{Best signal, compact heatmap, and pretrained heatmap results under the common protocol. The selected model may differ across metrics.}
\label{fig:signal}
\end{figure*}

Taken together, the heatmap models are competitive with direct signal processing. Their best EER values closely approach the strongest signal results on PTB and ECG-ID, while ConvNeXt-Tiny gives the highest PTB Rank-1 and DeiT-Base gives the highest PTB Rank-5 in the complete pretrained screen. On MIMIC-DEMO, DeiT-Base is best on four metrics and tied for best on Rank-5. Signal methods remain strongest for strict-FMR verification and Rank-1 on ECG-ID.

\subsection{Lead-channel scoring is cohort- and task-dependent}
\begin{figure*}[t]
\centering
\includegraphics[width=0.92\textwidth]{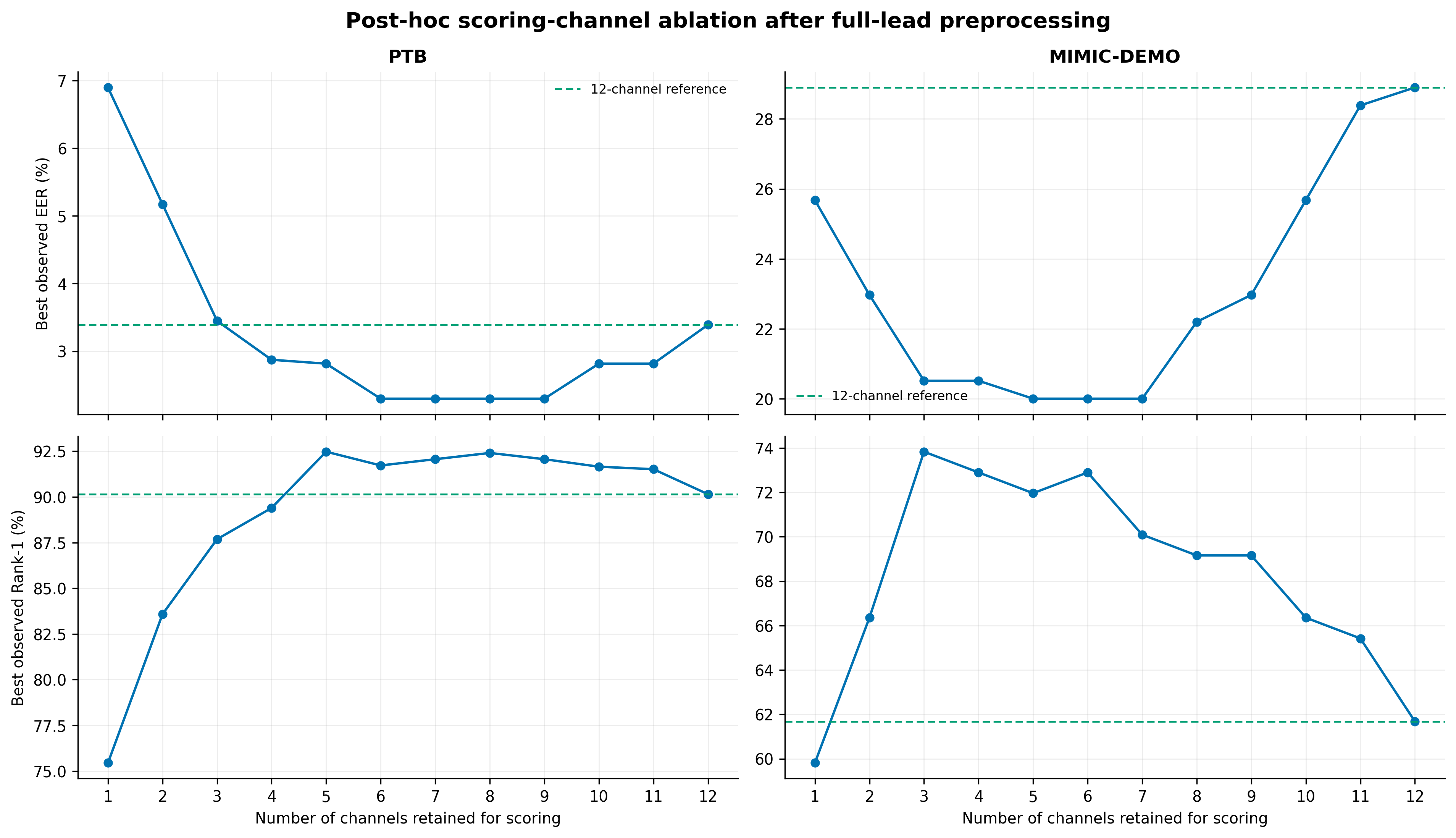}
\caption{Exhaustive scoring-channel ablation after full-lead preprocessing. Curves show the best EER and Rank-1 at each subset size. Dashed lines show the 12-channel reference.}
\label{fig:subsets}
\end{figure*}

The lead analysis uses summary-segment cosine similarity, with channels retained only after R-peak localization and representative-beat selection. Table~\ref{tab:leads} gives the one-channel and 12-channel references, while Figure~\ref{fig:subsets} summarizes all 4,095 scoring subsets in each multilead dataset.

PTB benefits from multiple scoring channels. V1 gives the lowest one-channel EER at 6.90\%, compared with 3.39\% for all 12 channels, while Lead II gives the highest one-channel Rank-1 at 75.44\%, compared with 90.15\% for the complete summary. A three-channel subset, aVL+V3+V5, reaches 3.45\% EER, and III+aVR+V3+V5 reaches 89.40\% Rank-1 with four channels. The lowest PTB EER is 2.30\% and occurs with subsets containing six to nine channels, whereas the highest Rank-1 is 92.48\% for III+aVR+V2+V3+V5.

MIMIC-DEMO follows a different pattern. Lead III alone gives 25.68\% EER, compared with 28.90\% for all channels, while aVF reaches 59.81\% Rank-1, compared with 61.68\% for the complete summary. Several subsets of five to seven channels obtain the lowest EER of 20.00\%, and the three-channel combination II+III+V1 gives the highest Rank-1 at 73.83\%. Verification and identification therefore favour different channel combinations.

\begin{table*}[t]
\centering
\caption{Lead-channel scoring ablation with summary cosine after full-lead preprocessing. The final columns use tolerances of one EER point and two Rank-1 points relative to the 12-channel reference.}
\label{tab:leads}
\small
\resizebox{\textwidth}{!}{%
\begin{tabular}{lccccc}
\toprule
Dataset & Best one-channel EER & Best one-channel R1 & 12-channel EER/R1 & Min. channels near 12L EER & Min. channels near 12L R1 \\
\midrule
PTB & V1 (6.90) & II (75.44) & 3.39 / 90.15 & 3 (aVL+V3+V5) & 4 (III+aVR+V3+V5) \\
MIMIC-DEMO & III (25.68) & aVF (59.81) & 28.90 / 61.68 & 1 (III) & 1 (aVF) \\
\bottomrule
\end{tabular}%
}
\end{table*}

The complete rankings indicate limited portability between datasets. Across all composition-matched subsets, Spearman correlation is $\rho=0.28$ for EER and $\rho=0.52$ for Rank-1. The size-stratified analysis in Appendix~\ref{app:concordance} gives a median concordance of $\rho=0.18$ for EER and $\rho=0.45$ for Rank-1 across $k=1,\ldots,11$, showing that verification-optimal combinations differ more strongly than identification-optimal combinations. Appendix~\ref{app:subsets} lists the best subset at every channel count. These results confirm that channel selection should reflect both the acquisition setting and the biometric objective.

\subsection{Limited average sensitivity to synthetic image degradation}
The degradation experiment evaluates Compact-TransMIL and ZACH-ViT without retraining, comparing clean performance with the average over eight image degradations in Table~\ref{tab:degradation}.

Mean EER changes remain small: +0.07 and +0.81 points for Compact-TransMIL and ZACH-ViT on PTB, +0.20 and +0.34 points on ECG-ID, and +0.57 and $-0.36$ points on MIMIC-DEMO. The small negative change in the final condition reflects score variation rather than a benefit from corruption.

Identification is slightly more sensitive, with mean Rank-1 decreases of 2.57 and 3.45 points on PTB, 1.01 and 1.87 points on ECG-ID, and 1.82 and 1.68 points on MIMIC-DEMO. Complete condition-level results are included in the supplementary data. Overall, both compact models remain stable under the tested noise, blur, compression, and resolution changes.

\begin{landscape}
\begin{table}[p]
\centering
\caption{Heatmap degradation stress test. Corrupted values first average the eight degradation conditions within each seed and then report mean $\pm$ sample SD over five seeds. EER, FNMR, and Rank-1 are percentages.}
\label{tab:degradation}
\scriptsize
\begin{tabular}{llcccccc}
\toprule
Dataset & Model & Clean EER & Corr. EER & Clean FNMR@$\leq$10 & Corr. FNMR@$\leq$10 & Clean R1 & Corr. R1 \\
\midrule
ECG-ID & Compact-TransMIL & 6.86 $\pm$ 2.06 & 7.06 $\pm$ 1.85 & 5.09 $\pm$ 3.73 & 5.27 $\pm$ 3.26 & 80.00 $\pm$ 2.52 & 78.99 $\pm$ 3.19 \\
ECG-ID & ZACH-ViT & 8.07 $\pm$ 0.96 & 8.41 $\pm$ 0.93 & 4.00 $\pm$ 2.99 & 6.95 $\pm$ 2.38 & 80.83 $\pm$ 1.16 & 78.96 $\pm$ 2.04 \\
MIMIC-DEMO & Compact-TransMIL & 24.54 $\pm$ 2.47 & 25.11 $\pm$ 2.75 & 40.57 $\pm$ 9.35 & 41.21 $\pm$ 5.40 & 61.31 $\pm$ 4.56 & 59.49 $\pm$ 5.16 \\
MIMIC-DEMO & ZACH-ViT & 27.95 $\pm$ 4.74 & 27.59 $\pm$ 5.61 & 38.29 $\pm$ 10.42 & 42.00 $\pm$ 10.04 & 65.23 $\pm$ 3.40 & 63.55 $\pm$ 1.92 \\
PTB & Compact-TransMIL & 6.33 $\pm$ 0.91 & 6.40 $\pm$ 0.46 & 5.06 $\pm$ 1.18 & 4.99 $\pm$ 0.82 & 73.82 $\pm$ 4.77 & 71.25 $\pm$ 3.16 \\
PTB & ZACH-ViT & 6.78 $\pm$ 0.49 & 7.60 $\pm$ 0.56 & 5.40 $\pm$ 1.04 & 6.08 $\pm$ 0.93 & 69.79 $\pm$ 2.17 & 66.34 $\pm$ 0.78 \\
\bottomrule
\end{tabular}
\end{table}
\end{landscape}

\begin{figure*}[t]
\centering
\includegraphics[width=0.94\textwidth]{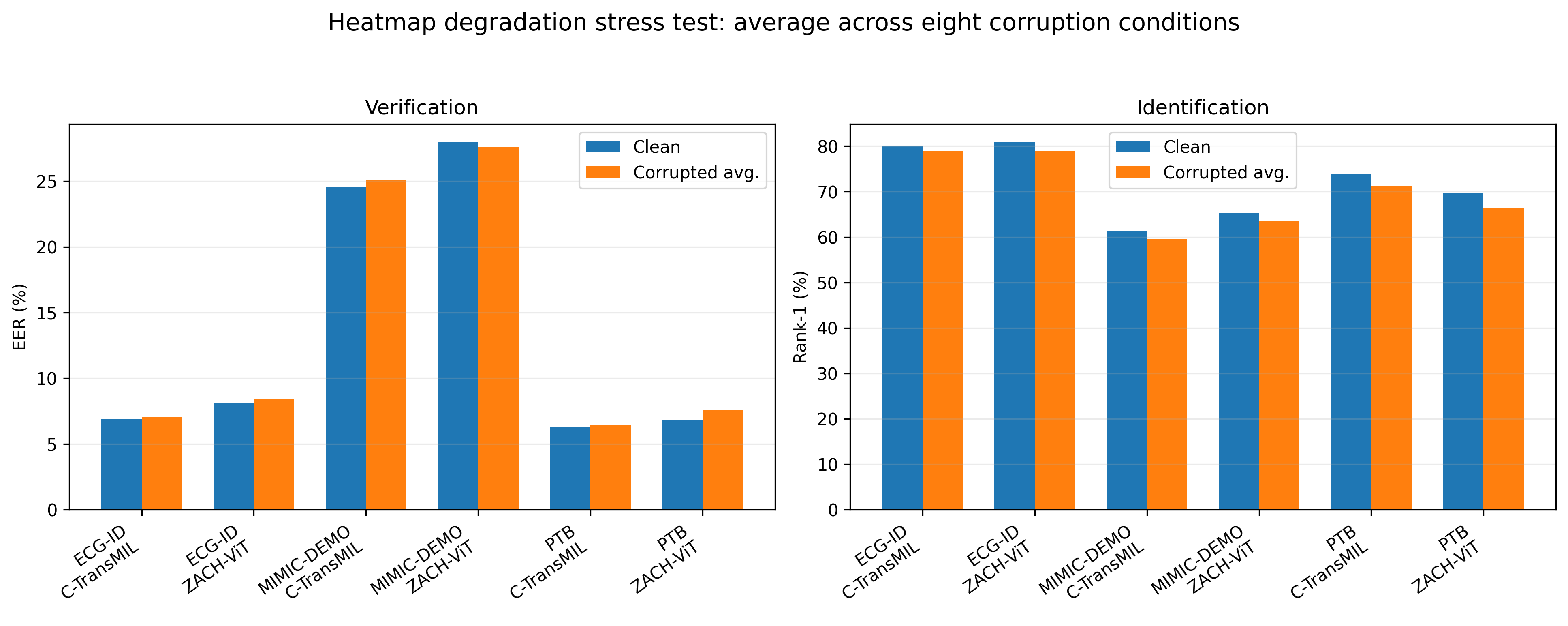}
\caption{Clean and corrupted-average performance under eight image-degradation conditions.}
\label{fig:degradation}
\end{figure*}

\section{Discussion}

\subsection{Representative morphology as an image}
The proposed representation is not simply a different visualization of an ECG waveform. It begins with an ECGXtractor-style procedure that selects five central beats from every block of ten and averages them into a representative matrix that preserves cardiac time, amplitude, and acquired leads. This ECG-informed summary provides the common physiological basis for both image representations.

Since the trace and heatmap use the same matrix, heartbeat selection, alignment, normalization, and numerical samples remain fixed while only the image encoding changes. Heatmaps improve both FNMR measures and both identification ranks in all 15 compact model-dataset comparisons, and mean EER improves in 14 of them. The consistency of these gains across architectures and cohorts supports a representation-level effect rather than an advantage restricted to a particular model.

ECG-ID provides an important control because its heatmap contains a single physiological lead and therefore has no inter-lead structure. Even in this setting, ZACH-ViT reduces EER from 22.75\% to 8.07\% and increases Rank-1 from 36.11\% to 80.83\%, with Compact-TransMIL showing a similarly large EER improvement. Dense amplitude encoding can therefore benefit recognition independently of the multilead image structure, giving convolution and attention direct access to the complete morphology rather than to a sparse line drawing.

This matched construction defines the central scientific contribution. Although earlier studies have used ECG images, colour maps, CNNs, and vision transformers, our method converts an ECGXtractor-style representative morphology into one dense time-by-lead image and compares it with a conventional trace derived from the identical numerical source. The experiment therefore measures how the image representation itself affects ECG biometric recognition.

\subsection{Visual pretraining and model size}
The pretrained benchmark extends the representation analysis to larger image encoders. On PTB and ECG-ID, the lowest EER is obtained by ConvNeXt-Tiny, whereas DeiT-Base is strongest on MIMIC-DEMO, and performance does not improve monotonically with the number of parameters. Architecture and initialization are consequently more informative than model size alone.

The matched ablations reveal a clear transfer effect on the multilead datasets, where ImageNet initialization substantially improves both ConvNeXt-Tiny and DeiT-Base. Its influence on ECG-ID is smaller and varies across metrics. One plausible interpretation is that generic visual features are particularly useful when the heatmap contains richer joint structure across cardiac time and multiple leads, although the present experiments do not establish this mechanism causally.

The compact models remain competitive even when compared with much larger architectures trained from scratch. CCT and Compact-TransMIL outperform both large scratch models in MIMIC-DEMO EER, while ZACH-ViT nearly matches scratch DeiT-Base in Rank-1 and exceeds it on ECG-ID. Achieving these results with approximately 0.25M parameters gives ZACH-ViT a favourable balance between accuracy and model footprint. The compact and large branches use different training schedules, however, so this comparison characterizes the achieved accuracy-size trade-off rather than controlled scaling.

\subsection{Comparison with signal-domain methods}
The strongest heatmap models are competitive with the direct signal baselines. ConvNeXt-Tiny approaches the best signal EER on PTB and ECG-ID while providing the highest PTB Rank-1, and DeiT-Base gives the highest PTB Rank-5 in the complete pretrained screen. On MIMIC-DEMO, DeiT-Base is best on four metrics and tied for best on Rank-5. Direct signal methods nevertheless remain strongest for strict-FMR verification and Rank-1 on ECG-ID.

These findings show that rasterization does not impose an inherent performance ceiling when the image is constructed from a stable and discriminative ECG morphology. The heatmap can preserve information required for verification and identification while providing access to mature computer vision architectures and pretrained weights. Direct signal models remain an important alternative when waveform-specific processing is preferred, and the two representations should therefore be viewed as complementary system designs rather than mutually exclusive solutions.

\subsection{Lead-channel contribution}
The exhaustive analysis shows that channel utility depends on both the dataset and the biometric task. PTB benefits from several scoring channels, although three channels nearly match the 12-channel EER and four channels nearly match its Rank-1. MIMIC-DEMO follows a different pattern, as selected single channels can match or improve the complete summary-cosine result.

Verification and identification also favour different combinations because they use the resulting score distributions in different ways. The cross-dataset correlations further show that the strongest subsets are not fully portable. Channel selection should therefore be guided by the intended acquisition setting and operating objective rather than by a universal ranking of ECG leads.

\subsection{Image degradation and deployment}
Compact-TransMIL and ZACH-ViT remain stable under the tested image degradations, with small average EER changes under Gaussian noise, blur, JPEG compression, and resolution reduction. Rank-1 is slightly more sensitive but remains close to the corresponding clean result. A related iris PAD study using Patch-ABMIL, Compact-TransMIL, and ZACH-ViT also found architecture-dependent differences under Gaussian noise, blur, and JPEG compression \cite{angelakis2026irispad}. Its unknown-PAI results further showed that relative robustness does not by itself establish deployment readiness.

The complete experiments reveal two useful deployment routes. Compact models support CPU training and a small memory footprint, whereas larger pretrained architectures provide stronger absolute performance on several datasets. Representative-morphology heatmaps accommodate both regimes within the same image-domain framework, allowing the model choice to follow the available computational resources and required operating point.

\subsection{Limitations}
The study has two main limitations. Although the biometric protocol uses distinct summary segments and excludes identity and sample leakage, it does not require gallery and probe samples to originate from separate recording sessions; long-term cross-session and cross-device performance therefore remains to be established. In addition, the lead analysis removes channels after multilead preprocessing, which makes it a measure of scoring-channel contribution rather than a prospective validation of reduced-sensor acquisition.

\section{Conclusion}
We introduced representative-morphology heatmaps for ECG biometric verification and identification. Starting from ECGXtractor-style beat selection, the method converts a $400\times L$ summary matrix into a dense time-by-lead image and compares it with a conventional trace containing the same physiological samples. This matched design isolates the effect of image representation.

Heatmaps improve both FNMR operating points and both identification ranks in all 15 compact model-dataset comparisons, while mean EER improves in 14. ZACH-ViT offers a particularly favourable balance between accuracy and model size, remaining competitive with architectures that contain substantially more parameters.

ImageNet initialization provides clear gains on PTB and MIMIC-DEMO, although its influence on ECG-ID is mixed. ConvNeXt-Tiny reaches 2.43\% EER on PTB and 5.79\% on ECG-ID, while DeiT-Base reaches 14.92\% EER and 81.68\% Rank-1 on MIMIC-DEMO. The best heatmap models approach the strongest signal EER on PTB and ECG-ID; on MIMIC-DEMO, DeiT-Base is best on four metrics and tied for best on Rank-5.

The lead-channel analysis provides a complementary result, showing that useful subsets depend on the dataset and biometric objective. Overall, representative-morphology heatmaps form a practical bridge between ECG signal processing and computer vision, supporting compact CPU models and larger pretrained architectures within the same representation.

\section*{Data and Reproducibility Statement}
PTB, ECG-ID, and MIMIC-IV-ECG-DEMO are publicly available under their respective access terms, and no source ECG recordings are redistributed with this manuscript.

The reproducibility package contains the manuscript source, figure code, protocol summaries, and seed-level results, including all operating points, lead-channel tables, degradation results, and pretrained model results. It also documents the ConvNeXt-Tiny and DeiT-Base initialization experiments, while provenance records identify the source archives and their SHA-256 hashes.

All stochastic experiments use seeds $\{3,5,7,11,13\}$, and the biometric protocol uses seed 3 to keep the verification comparisons and gallery-probe assignments identical across models and training seeds. Code for preprocessing, rendering, training, and biometric evaluation will be released upon acceptance; dataset access remains governed by the original providers.

\appendix

\begin{landscape}
\section{Complete matched trace-heatmap operating points}
\label{app:representation-full}
\scriptsize
\begin{longtable}{lllccccc}
\caption{Complete compact-model operating points for the matched representation experiment. Values are percentages, mean $\pm$ sample SD over five seeds. Trace and heatmap rows use the same ECG summaries, architecture, training budget, subject partition, and fixed biometric trials.}\\
\toprule
Dataset & Model & Representation & EER & FNMR@$\leq$10 & FNMR@$\leq$1 & Rank-1 & Rank-5 \\
\midrule
\endfirsthead
\toprule
Dataset & Model & Representation & EER & FNMR@$\leq$10 & FNMR@$\leq$1 & Rank-1 & Rank-5 \\
\midrule
\endhead
PTB & ABMIL & Trace & $9.56\pm2.42$ & $10.00\pm3.87$ & $23.91\pm5.05$ & $60.41\pm4.66$ & $77.59\pm4.99$ \\
PTB & ABMIL & Heatmap & $8.49\pm1.78$ & $7.93\pm2.38$ & $18.05\pm2.71$ & $65.27\pm4.05$ & $84.72\pm3.92$ \\
PTB & Patch-ABMIL & Trace & $20.69\pm2.77$ & $27.59\pm5.22$ & $51.61\pm8.82$ & $34.95\pm9.80$ & $53.01\pm10.17$ \\
PTB & Patch-ABMIL & Heatmap & $10.57\pm0.87$ & $11.03\pm1.43$ & $25.40\pm2.77$ & $55.77\pm1.72$ & $75.35\pm1.72$ \\
PTB & Compact-TransMIL & Trace & $10.74\pm1.13$ & $11.95\pm2.17$ & $29.89\pm6.84$ & $53.30\pm5.93$ & $73.78\pm5.17$ \\
PTB & Compact-TransMIL & Heatmap & $6.33\pm0.91$ & $5.06\pm1.18$ & $13.56\pm4.06$ & $73.82\pm4.77$ & $88.41\pm2.69$ \\
PTB & CCT-2/3x2 & Trace & $7.91\pm2.63$ & $7.59\pm4.75$ & $20.92\pm8.68$ & $62.12\pm11.86$ & $81.12\pm6.72$ \\
PTB & CCT-2/3x2 & Heatmap & $5.54\pm1.20$ & $4.25\pm1.19$ & $12.41\pm1.32$ & $74.05\pm3.67$ & $87.98\pm3.88$ \\
PTB & ZACH-ViT & Trace & $14.33\pm3.26$ & $16.90\pm5.17$ & $36.32\pm5.44$ & $48.24\pm4.36$ & $66.69\pm7.27$ \\
PTB & ZACH-ViT & Heatmap & $6.78\pm0.49$ & $5.40\pm1.04$ & $14.02\pm1.04$ & $69.79\pm2.17$ & $85.14\pm4.92$ \\
ECG-ID & ABMIL & Trace & $25.11\pm5.91$ & $47.27\pm10.20$ & $81.45\pm6.08$ & $44.44\pm18.89$ & $73.89\pm17.82$ \\
ECG-ID & ABMIL & Heatmap & $17.93\pm1.37$ & $25.09\pm3.25$ & $60.73\pm11.54$ & $71.11\pm6.16$ & $93.61\pm0.76$ \\
ECG-ID & Patch-ABMIL & Trace & $30.00\pm3.82$ & $62.18\pm12.69$ & $88.36\pm10.33$ & $27.22\pm2.88$ & $66.39\pm1.52$ \\
ECG-ID & Patch-ABMIL & Heatmap & $13.07\pm1.45$ & $16.73\pm3.25$ & $49.45\pm2.37$ & $56.94\pm1.96$ & $87.78\pm3.46$ \\
ECG-ID & Compact-TransMIL & Trace & $23.68\pm4.10$ & $44.36\pm8.29$ & $88.36\pm5.84$ & $40.28\pm11.74$ & $71.94\pm8.06$ \\
ECG-ID & Compact-TransMIL & Heatmap & $6.86\pm2.06$ & $5.09\pm3.73$ & $32.73\pm11.13$ & $80.00\pm2.52$ & $92.50\pm1.58$ \\
ECG-ID & CCT-2/3x2 & Trace & $24.14\pm7.19$ & $44.36\pm22.44$ & $74.18\pm15.02$ & $55.56\pm12.88$ & $85.83\pm12.71$ \\
ECG-ID & CCT-2/3x2 & Heatmap & $9.57\pm2.25$ & $9.09\pm3.64$ & $33.45\pm9.15$ & $73.06\pm5.25$ & $93.89\pm2.88$ \\
ECG-ID & ZACH-ViT & Trace & $22.75\pm3.66$ & $45.09\pm10.48$ & $87.64\pm10.86$ & $36.11\pm6.05$ & $73.89\pm10.27$ \\
ECG-ID & ZACH-ViT & Heatmap & $8.07\pm0.96$ & $4.00\pm2.99$ & $40.36\pm9.48$ & $80.83\pm1.16$ & $97.22\pm3.26$ \\
MIMIC-DEMO & ABMIL & Trace & $33.88\pm6.19$ & $69.71\pm12.55$ & $93.71\pm7.11$ & $28.22\pm5.62$ & $65.05\pm12.30$ \\
MIMIC-DEMO & ABMIL & Heatmap & $37.24\pm3.14$ & $61.14\pm9.17$ & $81.71\pm7.45$ & $40.93\pm6.39$ & $79.81\pm5.14$ \\
MIMIC-DEMO & Patch-ABMIL & Trace & $44.05\pm3.32$ & $77.14\pm8.33$ & $92.00\pm6.19$ & $22.99\pm6.79$ & $55.89\pm8.70$ \\
MIMIC-DEMO & Patch-ABMIL & Heatmap & $32.39\pm3.86$ & $54.29\pm4.52$ & $85.71\pm5.71$ & $43.36\pm4.26$ & $75.33\pm2.53$ \\
MIMIC-DEMO & Compact-TransMIL & Trace & $40.59\pm3.73$ & $76.57\pm6.19$ & $93.14\pm3.26$ & $21.12\pm4.84$ & $59.44\pm6.89$ \\
MIMIC-DEMO & Compact-TransMIL & Heatmap & $24.54\pm2.47$ & $40.57\pm9.35$ & $64.57\pm7.98$ & $61.31\pm4.56$ & $86.92\pm2.88$ \\
MIMIC-DEMO & CCT-2/3x2 & Trace & $31.66\pm5.21$ & $60.00\pm5.35$ & $88.57\pm2.02$ & $43.93\pm4.67$ & $72.34\pm4.21$ \\
MIMIC-DEMO & CCT-2/3x2 & Heatmap & $24.49\pm4.64$ & $44.00\pm7.45$ & $71.43\pm10.88$ & $58.69\pm9.49$ & $86.17\pm5.22$ \\
MIMIC-DEMO & ZACH-ViT & Trace & $44.54\pm6.28$ & $80.00\pm11.07$ & $96.00\pm4.33$ & $20.93\pm10.08$ & $54.58\pm15.28$ \\
MIMIC-DEMO & ZACH-ViT & Heatmap & $27.95\pm4.74$ & $38.29\pm10.42$ & $72.57\pm5.57$ & $65.23\pm3.40$ & $86.17\pm2.91$ \\
\bottomrule
\end{longtable}
\end{landscape}

\begin{landscape}
\section{Complete pretrained heatmap benchmark}
\label{app:pretrained-full}
\scriptsize
\begin{longtable}{llccccc}
\caption{Complete operating points for the six-model ImageNet-pretrained heatmap benchmark. Values are percentages, mean $\pm$ sample SD over five seeds. Every run uses full fine-tuning for 23 epochs and the same fixed biometric protocol.}\\
\toprule
Dataset & Model & EER & FNMR@$\leq$10 & FNMR@$\leq$1 & Rank-1 & Rank-5 \\
\midrule
\endfirsthead
\toprule
Dataset & Model & EER & FNMR@$\leq$10 & FNMR@$\leq$1 & Rank-1 & Rank-5 \\
\midrule
\endhead
PTB & ConvNeXt-Tiny & $2.43\pm0.50$ & $1.84\pm0.94$ & $3.79\pm0.31$ & $93.72\pm1.44$ & $98.11\pm0.84$ \\
PTB & EfficientNet-B0 & $2.87\pm0.81$ & $2.07\pm0.87$ & $5.86\pm1.37$ & $86.48\pm1.87$ & $96.24\pm1.59$ \\
PTB & DeiT-Small & $3.10\pm1.16$ & $1.49\pm0.87$ & $6.32\pm2.03$ & $88.15\pm2.41$ & $96.16\pm1.66$ \\
PTB & DeiT-Base & $3.13\pm0.85$ & $1.95\pm0.51$ & $5.06\pm0.75$ & $90.56\pm2.01$ & $98.18\pm0.41$ \\
PTB & DeiT-Tiny & $3.68\pm1.20$ & $2.41\pm1.25$ & $6.55\pm1.75$ & $87.03\pm1.81$ & $95.21\pm0.59$ \\
PTB & ResNet-18 & $3.79\pm0.66$ & $2.30\pm1.08$ & $7.24\pm2.98$ & $89.40\pm2.69$ & $97.80\pm0.27$ \\
ECG-ID & ConvNeXt-Tiny & $5.79\pm0.89$ & $2.18\pm1.52$ & $25.82\pm4.71$ & $84.72\pm2.20$ & $95.56\pm2.28$ \\
ECG-ID & EfficientNet-B0 & $6.25\pm0.90$ & $2.91\pm1.63$ & $33.82\pm19.14$ & $81.94\pm2.20$ & $96.11\pm2.48$ \\
ECG-ID & DeiT-Base & $6.35\pm0.97$ & $4.00\pm1.52$ & $34.55\pm3.86$ & $80.56\pm4.81$ & $97.22\pm3.26$ \\
ECG-ID & DeiT-Tiny & $7.29\pm1.08$ & $4.73\pm1.63$ & $34.55\pm9.09$ & $79.44\pm3.01$ & $96.67\pm3.20$ \\
ECG-ID & DeiT-Small & $7.61\pm1.48$ & $2.91\pm2.07$ & $29.09\pm6.68$ & $82.78\pm2.11$ & $98.61\pm1.70$ \\
ECG-ID & ResNet-18 & $10.82\pm2.12$ & $11.64\pm4.19$ & $36.00\pm7.20$ & $77.78\pm5.29$ & $93.33\pm2.67$ \\
MIMIC-DEMO & DeiT-Base & $14.92\pm3.08$ & $20.00\pm5.35$ & $54.86\pm14.76$ & $81.68\pm2.77$ & $96.64\pm1.07$ \\
MIMIC-DEMO & DeiT-Tiny & $19.36\pm2.55$ & $29.71\pm9.17$ & $53.14\pm7.98$ & $78.13\pm2.35$ & $94.39\pm1.62$ \\
MIMIC-DEMO & ConvNeXt-Tiny & $19.41\pm1.33$ & $28.00\pm4.69$ & $64.57\pm6.58$ & $78.69\pm5.14$ & $96.64\pm2.44$ \\
MIMIC-DEMO & DeiT-Small & $19.56\pm2.50$ & $30.29\pm5.92$ & $53.14\pm8.23$ & $77.20\pm4.00$ & $95.14\pm2.33$ \\
MIMIC-DEMO & ResNet-18 & $22.97\pm3.67$ & $36.00\pm5.57$ & $65.14\pm7.67$ & $68.22\pm5.84$ & $91.21\pm2.15$ \\
MIMIC-DEMO & EfficientNet-B0 & $25.29\pm2.72$ & $40.57\pm5.50$ & $62.86\pm13.85$ & $76.45\pm2.42$ & $90.47\pm1.80$ \\
\bottomrule
\end{longtable}
\end{landscape}

\begin{landscape}
\section{Complete matched initialization ablation}
\label{app:initialization-full}
\scriptsize
\begin{longtable}{lllccccc}
\caption{Complete operating points for the matched 23-epoch initialization ablation. All values are percentages, mean $\pm$ sample SD over five training seeds. Within each model-dataset pair only initialization changes.}\\
\toprule
Model & Dataset & Initialization & EER & FNMR@$\leq$10 & FNMR@$\leq$1 & Rank-1 & Rank-5 \\
\midrule
\endfirsthead
\toprule
Model & Dataset & Initialization & EER & FNMR@$\leq$10 & FNMR@$\leq$1 & Rank-1 & Rank-5 \\
\midrule
\endhead
ConvNeXt-Tiny & PTB & Scratch & $5.17\pm0.57$ & $4.02\pm0.91$ & $11.03\pm2.09$ & $78.92\pm2.53$ & $92.24\pm1.02$ \\
ConvNeXt-Tiny & PTB & ImageNet-1K & $2.43\pm0.50$ & $1.84\pm0.94$ & $3.79\pm0.31$ & $93.72\pm1.44$ & $98.11\pm0.84$ \\
ConvNeXt-Tiny & ECG-ID & Scratch & $6.64\pm1.01$ & $4.00\pm1.52$ & $24.00\pm1.52$ & $88.06\pm1.24$ & $100.00\pm0.00$ \\
ConvNeXt-Tiny & ECG-ID & ImageNet-1K & $5.79\pm0.89$ & $2.18\pm1.52$ & $25.82\pm4.71$ & $84.72\pm2.20$ & $95.56\pm2.28$ \\
ConvNeXt-Tiny & MIMIC-DEMO & Scratch & $26.66\pm3.27$ & $53.14\pm6.26$ & $83.43\pm3.13$ & $47.85\pm6.14$ & $82.80\pm3.53$ \\
ConvNeXt-Tiny & MIMIC-DEMO & ImageNet-1K & $19.41\pm1.33$ & $28.00\pm4.69$ & $64.57\pm6.58$ & $78.69\pm5.14$ & $96.64\pm2.44$ \\
DeiT-Base & PTB & Scratch & $5.85\pm0.94$ & $5.17\pm1.08$ & $10.80\pm1.37$ & $77.21\pm1.12$ & $89.86\pm0.86$ \\
DeiT-Base & PTB & ImageNet-1K & $3.13\pm0.85$ & $1.95\pm0.51$ & $5.06\pm0.75$ & $90.56\pm2.01$ & $98.18\pm0.41$ \\
DeiT-Base & ECG-ID & Scratch & $6.00\pm0.91$ & $2.91\pm2.07$ & $32.36\pm7.20$ & $79.72\pm0.76$ & $95.83\pm1.39$ \\
DeiT-Base & ECG-ID & ImageNet-1K & $6.35\pm0.97$ & $4.00\pm1.52$ & $34.55\pm3.86$ & $80.56\pm4.81$ & $97.22\pm3.26$ \\
DeiT-Base & MIMIC-DEMO & Scratch & $26.22\pm5.51$ & $43.43\pm5.11$ & $60.57\pm7.40$ & $65.42\pm1.14$ & $87.66\pm4.15$ \\
DeiT-Base & MIMIC-DEMO & ImageNet-1K & $14.92\pm3.08$ & $20.00\pm5.35$ & $54.86\pm14.76$ & $81.68\pm2.77$ & $96.64\pm1.07$ \\
\bottomrule
\end{longtable}
\end{landscape}

\begin{landscape}
\section{Best exhaustive subset at each lead count}
\label{app:subsets}
\scriptsize
\begin{longtable}{llp{4.8cm}ccp{5cm}cc}
\caption{Exact manuscript-selected subset at every lead count. Verification rows minimize EER; identification rows maximize Rank-1 after exact recomputation of near-tied candidates.}\\
\toprule
Dataset & $k$ & Best verification subset & EER & FNMR@$\leq$10 & Best identification subset & R1 & R5 \\
\midrule
\endfirsthead
\toprule
Dataset & $k$ & Best verification subset & EER & FNMR@$\leq$10 & Best identification subset & R1 & R5 \\
\midrule
\endhead
PTB & 1 & V1 & 6.90 & 6.32 & II & 75.44 & 86.25 \\
PTB & 2 & aVF+\allowbreak{}V2 & 5.17 & 2.87 & II+\allowbreak{}V2 & 83.58 & 90.77 \\
PTB & 3 & aVL+\allowbreak{}V3+\allowbreak{}V5 & 3.45 & 2.87 & III+\allowbreak{}aVR+\allowbreak{}V5 & 87.69 & 93.78 \\
PTB & 4 & aVF+\allowbreak{}V1+\allowbreak{}V3+\allowbreak{}V5 & 2.87 & 2.30 & III+\allowbreak{}aVR+\allowbreak{}V3+\allowbreak{}V5 & 89.40 & 96.37 \\
PTB & 5 & aVR+\allowbreak{}aVL+\allowbreak{}aVF+\allowbreak{}V3+\allowbreak{}V6 & 2.82 & 2.30 & III+\allowbreak{}aVR+\allowbreak{}V2+\allowbreak{}V3+\allowbreak{}V5 & 92.48 & 96.99 \\
PTB & 6 & aVR+\allowbreak{}aVL+\allowbreak{}aVF+\allowbreak{}V1+\allowbreak{}V3+\allowbreak{}V5 & 2.30 & 2.30 & I+\allowbreak{}II+\allowbreak{}III+\allowbreak{}V1+\allowbreak{}V2+\allowbreak{}V6 & 91.72 & 96.65 \\
PTB & 7 & I+\allowbreak{}aVR+\allowbreak{}aVL+\allowbreak{}aVF+\allowbreak{}V1+\allowbreak{}V3+\allowbreak{}V5 & 2.30 & 2.30 & II+\allowbreak{}III+\allowbreak{}aVR+\allowbreak{}V1+\allowbreak{}V2+\allowbreak{}V3+\allowbreak{}V6 & 92.07 & 97.20 \\
PTB & 8 & I+\allowbreak{}aVR+\allowbreak{}aVL+\allowbreak{}aVF+\allowbreak{}V2+\allowbreak{}V3+\allowbreak{}V5+\allowbreak{}V6 & 2.30 & 2.30 & II+\allowbreak{}III+\allowbreak{}aVR+\allowbreak{}aVL+\allowbreak{}V1+\allowbreak{}V2+\allowbreak{}V3+\allowbreak{}V5 & 92.41 & 96.92 \\
PTB & 9 & I+\allowbreak{}II+\allowbreak{}aVR+\allowbreak{}aVL+\allowbreak{}aVF+\allowbreak{}V1+\allowbreak{}V2+\allowbreak{}V3+\allowbreak{}V5 & 2.30 & 2.30 & II+\allowbreak{}III+\allowbreak{}aVR+\allowbreak{}aVL+\allowbreak{}V1+\allowbreak{}V2+\allowbreak{}V3+\allowbreak{}V5+\allowbreak{}V6 & 92.07 & 97.13 \\
PTB & 10 & I+\allowbreak{}II+\allowbreak{}III+\allowbreak{}aVR+\allowbreak{}aVL+\allowbreak{}V1+\allowbreak{}V2+\allowbreak{}V3+\allowbreak{}V5+\allowbreak{}V6 & 2.82 & 2.30 & II+\allowbreak{}III+\allowbreak{}aVR+\allowbreak{}aVL+\allowbreak{}aVF+\allowbreak{}V1+\allowbreak{}V2+\allowbreak{}V3+\allowbreak{}V5+\allowbreak{}V6 & 91.66 & 96.99 \\
PTB & 11 & I+\allowbreak{}II+\allowbreak{}III+\allowbreak{}aVR+\allowbreak{}aVL+\allowbreak{}aVF+\allowbreak{}V1+\allowbreak{}V2+\allowbreak{}V3+\allowbreak{}V5+\allowbreak{}V6 & 2.82 & 2.30 & I+\allowbreak{}II+\allowbreak{}III+\allowbreak{}aVR+\allowbreak{}aVL+\allowbreak{}aVF+\allowbreak{}V1+\allowbreak{}V2+\allowbreak{}V3+\allowbreak{}V5+\allowbreak{}V6 & 91.52 & 97.33 \\
PTB & 12 & I+\allowbreak{}II+\allowbreak{}III+\allowbreak{}aVR+\allowbreak{}aVL+\allowbreak{}aVF+\allowbreak{}V1+\allowbreak{}V2+\allowbreak{}V3+\allowbreak{}V4+\allowbreak{}V5+\allowbreak{}V6 & 3.39 & 2.30 & I+\allowbreak{}II+\allowbreak{}III+\allowbreak{}aVR+\allowbreak{}aVL+\allowbreak{}aVF+\allowbreak{}V1+\allowbreak{}V2+\allowbreak{}V3+\allowbreak{}V4+\allowbreak{}V5+\allowbreak{}V6 & 90.15 & 96.72 \\
MIMIC-DEMO & 1 & III & 25.68 & 42.86 & aVF & 59.81 & 85.05 \\
MIMIC-DEMO & 2 & III+\allowbreak{}V4 & 22.97 & 40.00 & III+\allowbreak{}V1 & 66.36 & 92.52 \\
MIMIC-DEMO & 3 & III+\allowbreak{}aVL+\allowbreak{}V2 & 20.51 & 40.00 & II+\allowbreak{}III+\allowbreak{}V1 & 73.83 & 93.46 \\
MIMIC-DEMO & 4 & III+\allowbreak{}aVL+\allowbreak{}V2+\allowbreak{}V4 & 20.51 & 34.29 & II+\allowbreak{}III+\allowbreak{}aVL+\allowbreak{}V1 & 72.90 & 89.72 \\
MIMIC-DEMO & 5 & III+\allowbreak{}aVL+\allowbreak{}V1+\allowbreak{}V3+\allowbreak{}V4 & 20.00 & 34.29 & II+\allowbreak{}III+\allowbreak{}aVL+\allowbreak{}aVF+\allowbreak{}V1 & 71.96 & 92.52 \\
MIMIC-DEMO & 6 & III+\allowbreak{}aVL+\allowbreak{}aVF+\allowbreak{}V1+\allowbreak{}V2+\allowbreak{}V4 & 20.00 & 31.43 & II+\allowbreak{}III+\allowbreak{}aVF+\allowbreak{}V1+\allowbreak{}V2+\allowbreak{}V6 & 72.90 & 89.72 \\
MIMIC-DEMO & 7 & I+\allowbreak{}III+\allowbreak{}aVL+\allowbreak{}aVF+\allowbreak{}V2+\allowbreak{}V3+\allowbreak{}V4 & 20.00 & 34.29 & II+\allowbreak{}III+\allowbreak{}aVL+\allowbreak{}aVF+\allowbreak{}V1+\allowbreak{}V2+\allowbreak{}V6 & 70.09 & 91.59 \\
MIMIC-DEMO & 8 & I+\allowbreak{}III+\allowbreak{}aVL+\allowbreak{}aVF+\allowbreak{}V1+\allowbreak{}V2+\allowbreak{}V3+\allowbreak{}V4 & 22.20 & 37.14 & II+\allowbreak{}III+\allowbreak{}aVR+\allowbreak{}aVL+\allowbreak{}aVF+\allowbreak{}V1+\allowbreak{}V4+\allowbreak{}V5 & 69.16 & 84.11 \\
MIMIC-DEMO & 9 & I+\allowbreak{}III+\allowbreak{}aVL+\allowbreak{}aVF+\allowbreak{}V1+\allowbreak{}V2+\allowbreak{}V3+\allowbreak{}V4+\allowbreak{}V5 & 22.97 & 42.86 & I+\allowbreak{}II+\allowbreak{}III+\allowbreak{}aVL+\allowbreak{}aVF+\allowbreak{}V1+\allowbreak{}V2+\allowbreak{}V4+\allowbreak{}V5 & 69.16 & 82.24 \\
MIMIC-DEMO & 10 & I+\allowbreak{}II+\allowbreak{}III+\allowbreak{}aVL+\allowbreak{}aVF+\allowbreak{}V1+\allowbreak{}V2+\allowbreak{}V3+\allowbreak{}V4+\allowbreak{}V5 & 25.68 & 40.00 & II+\allowbreak{}III+\allowbreak{}aVR+\allowbreak{}aVL+\allowbreak{}aVF+\allowbreak{}V1+\allowbreak{}V2+\allowbreak{}V4+\allowbreak{}V5+\allowbreak{}V6 & 66.36 & 84.11 \\
MIMIC-DEMO & 11 & I+\allowbreak{}II+\allowbreak{}III+\allowbreak{}aVR+\allowbreak{}aVL+\allowbreak{}aVF+\allowbreak{}V2+\allowbreak{}V3+\allowbreak{}V4+\allowbreak{}V5+\allowbreak{}V6 & 28.39 & 45.71 & I+\allowbreak{}II+\allowbreak{}III+\allowbreak{}aVR+\allowbreak{}aVL+\allowbreak{}aVF+\allowbreak{}V1+\allowbreak{}V3+\allowbreak{}V4+\allowbreak{}V5+\allowbreak{}V6 & 65.42 & 81.31 \\
MIMIC-DEMO & 12 & I+\allowbreak{}II+\allowbreak{}III+\allowbreak{}aVR+\allowbreak{}aVL+\allowbreak{}aVF+\allowbreak{}V1+\allowbreak{}V2+\allowbreak{}V3+\allowbreak{}V4+\allowbreak{}V5+\allowbreak{}V6 & 28.90 & 37.14 & I+\allowbreak{}II+\allowbreak{}III+\allowbreak{}aVR+\allowbreak{}aVL+\allowbreak{}aVF+\allowbreak{}V1+\allowbreak{}V2+\allowbreak{}V3+\allowbreak{}V4+\allowbreak{}V5+\allowbreak{}V6 & 61.68 & 80.37 \\
\bottomrule
\end{longtable}
\end{landscape}

\section{Cross-dataset subset concordance}
\label{app:concordance}
\begin{table}[h]
\centering
\caption{Spearman rank concordance between PTB and MIMIC-DEMO for identical lead subsets at each subset size. Conventional $p$ values are nominal descriptors only because the exhaustive, overlapping subsets are not independent random observations.}
\small
\begin{tabular}{rrrrrr}
\toprule
$k$ & Number of subsets & $\rho$ EER & $p$ & $\rho$ Rank-1 & $p$ \\
\midrule
1 & 12 & -0.06 & 0.850 & 0.19 & 0.564 \\
2 & 66 & 0.07 & 0.557 & 0.32 & 0.009 \\
3 & 220 & 0.22 & <0.001 & 0.41 & <0.001 \\
4 & 495 & 0.32 & <0.001 & 0.45 & <0.001 \\
5 & 792 & 0.36 & <0.001 & 0.50 & <0.001 \\
6 & 924 & 0.33 & <0.001 & 0.51 & <0.001 \\
7 & 792 & 0.27 & <0.001 & 0.49 & <0.001 \\
8 & 495 & 0.18 & <0.001 & 0.42 & <0.001 \\
9 & 220 & -0.02 & 0.801 & 0.44 & <0.001 \\
10 & 66 & -0.02 & 0.860 & 0.60 & <0.001 \\
11 & 12 & -0.04 & 0.907 & 0.72 & 0.008 \\
\bottomrule
\end{tabular}
\end{table}

\bibliographystyle{unsrtnat_singlehyphen}
\bibliography{references}
\end{document}